\documentclass[11pt,a4paper]{article}
\usepackage{times,latexsym}
\usepackage{url}
\usepackage[T1]{fontenc}

\usepackage[acceptedWithA]{tacl2018v2}

\usepackage{xspace,mfirstuc,tabulary}

\newif\iftaclinstructions
\taclinstructionsfalse %
\iftaclinstructions
\renewcommand{\confidential}{}
\renewcommand{\anonsubtext}{(No author info supplied here, for consistency with
TACL-submission anonymization requirements)}
\newcommand{\instr}
\fi

\iftaclpubformat %

\else

\fi

\usepackage{times}
\usepackage{latexsym}
\usepackage{amsmath}
\usepackage{changepage}
\usepackage{graphicx}
\usepackage{colortbl}
\usepackage{subcaption}
\usepackage{float}
\usepackage{cleveref}
\usepackage{booktabs}
\usepackage{multirow}
\usepackage{makecell}
\usepackage{algorithm,algorithmic}
\usepackage{mathtools}
\usepackage[normalem]{ulem}
\usepackage{color,soul}
\usepackage{subcaption}
\usepackage{amsmath,amssymb}
\usepackage{bbold}
\usepackage{bm}
\usepackage{array}
\usepackage[normalem]{ulem}
\newcolumntype{?}{!{\vrule width 1pt}}
\useunder{\uline}{\ul}{}

\definecolor{celadon}{rgb}{0.67, 0.88, 0.69}
\definecolor{salmonpink}{rgb}{1.0, 0.57, 0.64}
\definecolor{lightpink}{rgb}{1.0, 0.71, 0.76}
\definecolor{beaublue}{rgb}{0.74, 0.83, 0.9}
\definecolor{indianyellow}{rgb}{0.89, 0.66, 0.34}
\definecolor{lightapricot}{rgb}{0.99, 0.84, 0.69}
\definecolor{brickred}{rgb}{0.8, 0.25, 0.33}
\definecolor{brightmaroon}{rgb}{0.76, 0.13, 0.28}

\newcommand{\hlp}[2][beaublue]{{\sethlcolor{#1}\hl{#2}} }
\newcommand{\hln}[2][lightapricot]{{\sethlcolor{#1}\hl{#2}} }
\newcommand{\hlnew}[2][lightpink]{{\sethlcolor{#1}\hl{#2}} }
\definecolor{lemonchiffon}{rgb}{1.0, 0.98, 0.8}

\usepackage{microtype}

\def\showcomments{1}
\if\showcomments1

\newcommand{\dd}[1]{\textcolor{red}{[  #1 -- Danish]}\typeout{#1}}
\newcommand{\wc}[1]{\textcolor{red}{[  #1 -- William]}\typeout{#1}}
\newcommand{\bd}[1]{\textcolor{orange}{[  #1 -- Bhuwan]}\typeout{#1}}
\newcommand{\gn}[1]{\textcolor{purple}{[  #1 -- GN]}\typeout{#1}}

\newcommand{\zl}[1]{\textcolor{blue}{[  #1 -- Zack]}\typeout{#1}}
\newcommand{\rb}[1]{\textcolor{teal}{[  #1 -- Rachit]}\typeout{#1}}
\else

\newcommand{\dd}[1]{}
\newcommand{\wc}[1]{}
\newcommand{\bd}[1]{}
\newcommand{\gn}[1]{}
\newcommand{\zl}[1]{}
\newcommand{\rb}[1]{}
\fi 

\title{ 
Evaluating Explanations: How much do explanations \\ from the teacher aid students? 
} %

\author{Danish Pruthi$^1$\thanks{\enskip Part of this work was done at Google.} \quad Rachit Bansal$^2$ \quad Bhuwan Dhingra$^3$ \quad Livio Baldini Soares$^3$ \\
\textbf{Michael Collins$^3$ \quad Zachary C. Lipton$^1$ \quad Graham Neubig$^1$ \quad William W. Cohen$^3$} \\
$^1$~Carnegie Mellon University \quad 
 $^2$~Delhi Technological University \\
 $^3$~Google Research \\
\texttt{\{ddanish, zlipton, gneubig\}@cs.cmu.edu, racbansa@gmail.com} \\
\texttt{\{bdhingra, liviobs, mjcollins, wcohen\}@google.com}}%

\begin{document}
\maketitle
\begin{abstract}
While many methods purport 
to \emph{explain} predictions 
by highlighting salient features, 
what aims these explanations serve
and how they ought to be evaluated
often go unstated.
In this work, we 
introduce a framework to
quantify the value of explanations 
via the accuracy gains
that they confer on a \emph{student} model
trained to simulate a \emph{teacher} model.
Crucially, the explanations are available
to the student during training,
but are not available at test time.
Compared to prior proposals,
our approach is less easily gamed,
enabling principled, automatic, model-agnostic 
evaluation of attributions.
Using our framework, 
we compare numerous attribution methods 
for text classification and question answering,
and observe quantitative differences 
that are consistent (to a moderate to high degree) 
across different student model architectures
and learning strategies.\footnote{Code for the evaluation protocol: \url{https://github.com/danishpruthi/evaluating-explanations}}

\end{abstract}
\section{Introduction}

The success of deep learning models, 
together with the difficulty 
of understanding how they work,
has inspired a subfield of research 
on \emph{explaining} predictions,
often by highlighting specific input features
deemed somehow \emph{important} to a prediction 
\cite{ribeiro2016should, sundararajan2017axiomatic, shrikumar2017learning}.
For instance, we might expect such a method
to highlight spans like ``poorly acted'' and ``slow-moving'' 
to explain a prediction of negative sentiment
for a given movie review.
However, there is little agreement in the literature
as to what constitutes a good explanation 
\cite{lipton2018mythos, jacovi2020aligning}.
Moreover, various popular methods
for generating such attributions
disagree considerably over 
which tokens to highlight
(Table~\ref{tab:overlap}).
With so many methods claimed 
to confer the same property 
while disagreeing so markedly,
one path forward is to develop
clear \emph{quantitative} criteria
for evaluating purported explanations at scale.

\begin{figure}[t]
    \centering
    \includegraphics[width=0.48\textwidth]{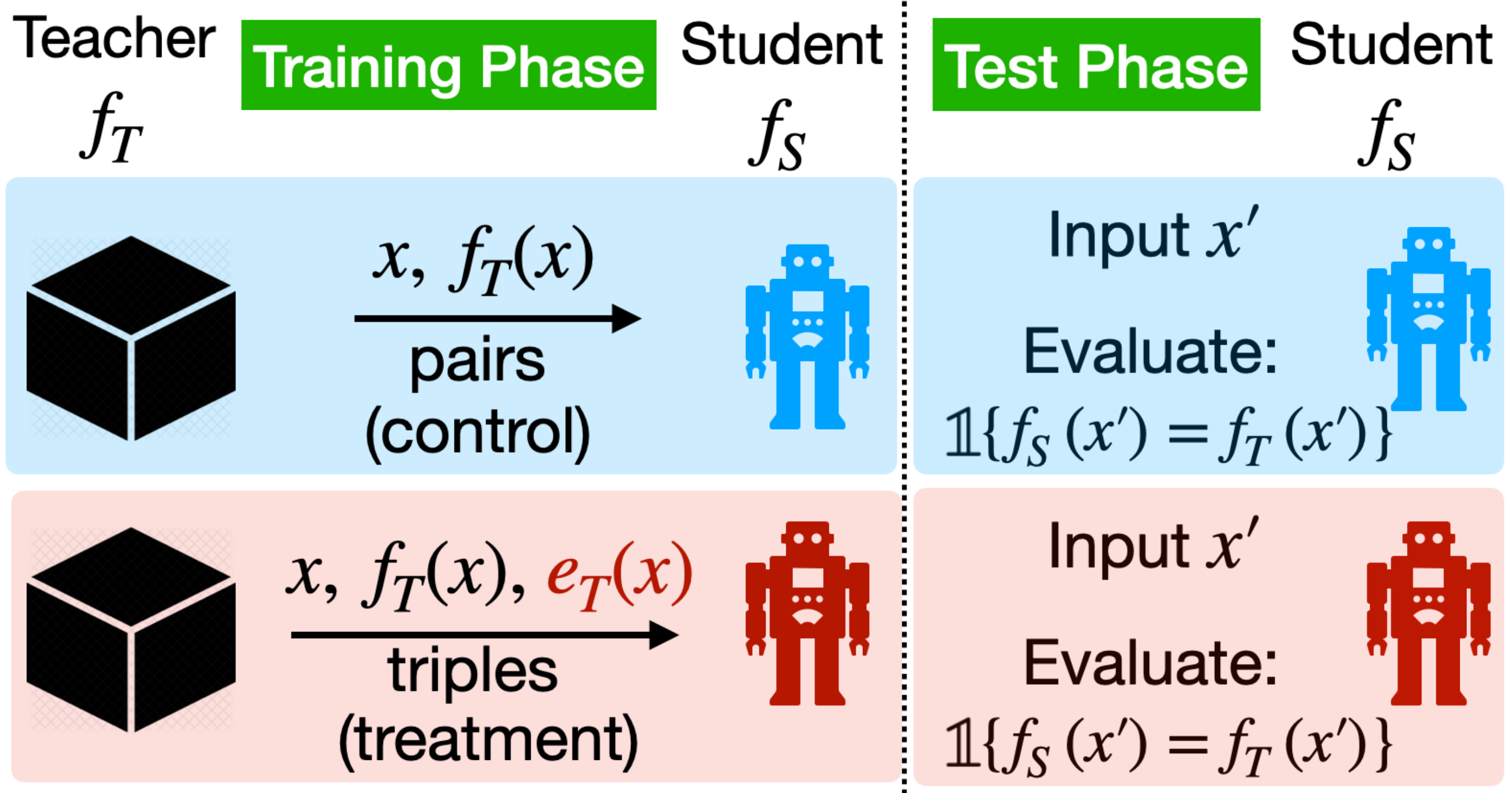}
    \caption{The proposed framework for quantifying explanation quality. 
    Student models learn to mimic the teacher, 
    with and without explanations 
    (provided as ``side information'' with each example). 
    Explanations are effective if they help students 
    to better approximate the teacher on unseen examples 
    \emph{for which explanations are not available}. 
    Students and teachers could be either models or people.
    }
    \label{fig:overview}
\end{figure}

The status quo for evaluating so-called
explanations skews qualitative---many 
proposed techniques are evaluated 
only via visual inspection of a few examples \cite{simonyan2013deep,sundararajan2017axiomatic,shrikumar2017learning}.
While several quantitative evaluation techniques
have recently been proposed, 
many of these are easily gamed
\citep{treviso2020towards,hase2020leakage}.\footnote{See \S\ref{sec:related_work} for a comprehensive discussion on existing metrics, 
and how they can be gamed by trivial strategies.}
Some depend upon the model outputs corresponding to deformed examples 
that lie outside the support of the training distribution
\cite{deyoung-etal-2020-eraser},
and a few validate explanations on specifically crafted tasks~\cite{poerner-etal-2018-evaluating}.

\begin{table*}[ht]
\centering
\footnotesize
\begin{tabular}{@{}l|c|c|c|c|c|c|c|c@{}}
\toprule
\multicolumn{1}{l}{ } &
  \multicolumn{1}{l}{Random} &
  \multicolumn{1}{l}{Grad Norm} &
  \multicolumn{1}{l}{Grad $\times$ Inp} &
  \multicolumn{1}{l}{LIME} &
  \multicolumn{1}{l}{DeepLIFT} &
  \multicolumn{1}{r}{\begin{tabular}[r]{@{}l@{}}Layer\\ Cond.\end{tabular}} & 
  \multicolumn{1}{r}{\begin{tabular}[r]{@{}l@{}}Integrated \\ Gradients\end{tabular}} &
  \multicolumn{1}{l}{Attention} \\  \midrule
Random &
  \cellcolor[HTML]{BCD4E6}1.00 &
  \cellcolor[HTML]{FDDDC0}0.10 &
  \cellcolor[HTML]{FDDDC0}0.10 &
  \cellcolor[HTML]{FDDDC0}0.10 &
  \cellcolor[HTML]{FDDDC0}0.10 &
  \cellcolor[HTML]{FDDDC0}0.10 &
  \cellcolor[HTML]{FDDDC0}0.10 &
  \cellcolor[HTML]{FDDDC0}0.10 \\ 
  \cline{2-9} 
Grad Norm &
  \cellcolor[HTML]{FDDDC0}0.10 &
  \cellcolor[HTML]{BCD4E6}1.00 &
  \cellcolor[HTML]{FEEBDB}0.27 &
  \cellcolor[HTML]{FDE0C5}0.13 &
  \cellcolor[HTML]{FDE7D3}0.22 &
  \cellcolor[HTML]{FDE5CF}0.19 &
  \cellcolor[HTML]{FDE7D2}0.22 &
  \cellcolor[HTML]{FEEEE0}0.30 \\ 
  \cline{2-9} 
Grad$\times$Inp &
  \cellcolor[HTML]{FDDDC0}0.10 &
  \cellcolor[HTML]{FEEBDB}0.27 &
  \cellcolor[HTML]{BCD4E6}1.00 &
  \cellcolor[HTML]{FDDEC2}0.11 &
  \cellcolor[HTML]{FEECDC}0.28 &
  \cellcolor[HTML]{FDE0C6}0.14 &
  \cellcolor[HTML]{FDE2C9}0.16 &
  \cellcolor[HTML]{FDE2CA}0.17 \\
  \cline{2-9} 
LIME &
  \cellcolor[HTML]{FDDDC0}0.10 &
  \cellcolor[HTML]{FDE0C5}0.13 &
  \cellcolor[HTML]{FDDEC2}0.11 &
  \cellcolor[HTML]{BCD4E6}1.00 &
  \cellcolor[HTML]{FDDDC1}0.11 &
  \cellcolor[HTML]{FDE2C9}0.16 &
  \cellcolor[HTML]{FDE2CA}0.16 &
  \cellcolor[HTML]{FDE1C8}0.15 \\
  \cline{2-9} 
DeepLIFT &
  \cellcolor[HTML]{FDDDC0}0.10 &
  \cellcolor[HTML]{FDE7D3}0.22 &
  \cellcolor[HTML]{FEECDC}0.28 &
  \cellcolor[HTML]{FDDDC1}0.11 &
  \cellcolor[HTML]{BCD4E6}1.00 &
  \cellcolor[HTML]{FDE3CC}0.18 &
  \cellcolor[HTML]{FDE8D4}0.23 &
  \cellcolor[HTML]{FDE9D7}0.25 \\
  \cline{2-9} 
Layer Cond. &
  \cellcolor[HTML]{FDDDC0}0.10 &
  \cellcolor[HTML]{FDE5CF}0.19 &
  \cellcolor[HTML]{FDE0C6}0.14 &
  \cellcolor[HTML]{FDE2C9}0.16 &
  \cellcolor[HTML]{FDE3CC}0.18 &
  \cellcolor[HTML]{BCD4E6}1.00 &
  \cellcolor[HTML]{FEFEFD}0.49 &
  \cellcolor[HTML]{FDE7D2}0.22 \\
  \cline{2-9} 
Integrated Gradients &
  \cellcolor[HTML]{FDDDC0}0.10 &
  \cellcolor[HTML]{FDE7D2}0.22 &
  \cellcolor[HTML]{FDE2C9}0.16 &
  \cellcolor[HTML]{FDE2CA}0.16 &
  \cellcolor[HTML]{FDE8D4}0.23 &
  \cellcolor[HTML]{FEFEFD}0.49 &
  \cellcolor[HTML]{BCD4E6}1.00 &
  \cellcolor[HTML]{FDE9D6}0.24 \\
  \cline{2-9} 
Attention &
  \cellcolor[HTML]{FDDDC0}0.10 &
  \cellcolor[HTML]{FEEEE0}0.30 &
  \cellcolor[HTML]{FDE2CA}0.17 &
  \cellcolor[HTML]{FDE1C8}0.15 &
  \cellcolor[HTML]{FDE9D7}0.25 &
  \cellcolor[HTML]{FDE7D2}0.22 &
  \cellcolor[HTML]{FDE9D6}0.24 &
  \cellcolor[HTML]{BCD4E6}1.00 \\ \bottomrule
\end{tabular}
\caption{Overlap among the top-$10\%$ tokens selected 
by different explanation techniques for sentiment analysis. 
In each row, for a given technique,
we tabulate the fraction of explanatory tokens 
that overlap with other explanations. 
Value of \hlp{1.0} implies perfect overlap 
and \hln{0.0} denotes no overlap.
}
\label{tab:overlap}
\end{table*}

In this work, we propose a new 
framework,  
where explanations are quantified 
by \emph{the degree to which 
they help a student model 
in learning to simulate 
the teacher on future examples} (Figure~\ref{fig:overview}). 
Our framework addresses a coherent goal,
is model-agnostic and broadly applicable across tasks,
and (when instantiated with models as students)
can easily be automated and scaled. 
Our method is inspired by argumentative models 
for justifying human reasoning,
which posit that the role of 
explanations is to communicate 
information about how decisions are made, 
and thus to enable a recipient 
to anticipate future decisions~\cite{mercier2017enigma}. 
Our framework is similar 
to human studies conducted by~\citet{hase-bansal-2020-evaluating},
who evaluate if explanations 
help predict model behavior.
However, here we focus on protocols 
that do not rely on human-subject experiments.

Using our framework, 
we conduct extensive experiments 
on two broad categories of NLP tasks: 
text classification and
question answering.
For classification tasks, 
we compare seven widely used input attribution techniques,
covering gradient-based methods~\cite{simonyan2013deep, sundararajan2017axiomatic}, perturbation-based techniques~\cite{ribeiro2016should}, 
attention-based explanations~\cite{bahdanau2014neural},
and other popular attributions~\cite{shrikumar2017learning, dhamdhere2018important}. 
These comparisons lead to observable 
quantitative differences---we find 
attention-based explanations 
and integrated gradients~\citep{sundararajan2017axiomatic} 
to be the most effective, 
and vanilla gradient-based saliency maps
and LIME to be the least effective.
Further, we observe moderate to high agreement 
among rankings obtained by varying 
student architectures and learning strategies in our framework.
For question answering, we validate 
the effectiveness of student learners 
on both human-produced explanations 
collected by~\citet{lamm2020qed}, 
and automatically generated explanations 
from a SpanBERT model~\cite{joshi2020spanbert}.

\section{Explanation as Communication}
\label{sec:exp-as-comm}

\subsection{An Illustrative Example}

In our framework, we view explanations 
as a communication channel 
between a teacher $T$ and a student $S$, 
whose purpose is to help $S$ 
to predict $T$'s outputs on a given input. 
As an example, consider the case 
of graduate admissions:
an aspirant submits their application $\bm{x}$ 
and subsequently the admission committee $T$
decides whether the candidate 
is to be accepted or not. %
The acceptance criterion, $f_T(\bm{x})$, 
represents a typical black box function---one 
that is of great interest to future aspirants.\footnote{Our 
illustrative example assumes 
that the admission decision 
depends solely upon the student application, 
and ignores how other competing applicants
might affect the outcome.}
To \emph{simulate} the admission criterion, 
a student $S$ might study profiles 
of several applicants from previous iterations, 
$\bm{x}_1, \dots, \bm{x}_n$,
and their admission outcomes 
$f_T(\bm{x}_1), \dots, f_T(\bm{x}_n)$.
Let $A(f_S, f_T)$ be 
the \emph{simulation accuracy},
i.e., the accuracy with which the student 
predicts the teacher's decisions 
on unseen future applications
(defined formally below in~\S\ref{subsec:quantification}).

Now suppose each previous admission outcome 
was supplemented with an additional 
explanation $\bm{e}_T(\bm{x})$, 
from the admission committee, 
intended to help $S$ understand 
the decisions made by $T$. %
Ideally, these explanations would 
enhance students' understanding 
about the admission process, and 
would help students simulate 
the admission decisions better,
leading to a higher accuracy.
We argue that the degree of improvement
in simulation accuracy 
is a quantitative indicator 
of the utility of the explanations. 
Note that generic %
explanations 
or explanations that simply encode the final decision (e.g.,~``We received far too many applications ...'') 
are unlikely to help students simulate $f_T(\bm{x})$, 
as they provide no \emph{additional} information.

\newcommand{\lrn}{\textit{learn}}
\newcommand{\loss}{{\ell}}

\subsection{Quantifying Explanations}
\label{subsec:quantification}

For concreteness, we assume a classification task, 
and for a teacher $T$, we let $f_T$ denote a model 
that computes the teacher's predictions. %
Let $S$ be a student (either human or a  machine),
then $T$ could teach $S$ to simulate $f_T$ by sampling $n$ examples, $\bm{x}_1,\dots,\bm{x}_n$ and sharing with $S$ 
a dataset $\hat{D}$ 
containing its associated  predictions 
$\{ (\bm{x}_1,\hat{y}_1),\ldots,(\bm{x}_n,\hat{y}_n)
\}$, where $\hat{y_i} = f_T(\bm{x_i})$, 
and $S$ could then learn some approximation of $f_T$ from this data: 
$$f_{S,\hat{D}} = \text{learn}(S, \hat{D})$$
Additionally, we assume that for a given teacher $T$, an explanation generation method can generate an explanation $\bm{e}_T(\bm{x})$ for any example $\bm{x}$ 
which is some side information that potentially helps $S$ in predicting $f_T(\bm{x})$. 
We use $\hat{E}$ to denote a dataset 
of explanation-augmented examples, i.e.,
$$\hat{E}=\{(\bm{x}_1, \bm{e}_T(\bm{x}_1),\hat{y}_1),\dots,(\bm{x}_n, \bm{e}_T(\bm{x}_n),\hat{y}_n)\}
$$
and the student learner can make use 
of this side information during training, 
to learn a classifier
$$f_{S,\hat{E}} = \text{learn}(S, \hat{E})$$

Note that none of the learning tasks discussed above 
involve the ``gold'' label $y$ for any instance $\bm{x}$, 
only the prediction $\hat{y}$ for $\bm{x}$, produced by the teacher.  
While the student $S$ can use the explanations for learning, 
all the classifiers $f_T$, $f_{S,\hat{D}}$, 
and $f_{S,\hat{E}}$ 
predict labels given only the input $\bm{x}$, 
without using the explanations, i.e.,
\emph{explanations are only available during training, not at test time.}

In our framework the benefit of explanations 
is measured by how much they help 
the student to simulate the teacher.
In particular, we quantify the ability 
of a student $f_S$ to simulate a teacher 
using the \emph{simulation accuracy}:
\begin{equation} \label{eq:explain-acc}
A(f_S, f_T) = \textrm{E}_{\bm{x}}\left[ 
 ~\mathbb{1}\{ f_{S}(\bm{x}) = f_T(\bm{x}) \} ~\right],
\end{equation}
where the expected agreement 
between student and teacher
is computed over test examples. 
Better explanations will lead to 
higher values of $A(f_{S,\hat{E}}, f_T)$
than the accuracy 
associated with learning to simulate 
the teacher without explanations, 
i.e. $A(f_{S,\hat{D}}, f_T)$.

So far, 
for a given teacher model, 
our criteria for explanation quality
depends upon the 
choice of the student model ($S$), its learning procedure and 
the number of examples used to train it ($n$).
To reduce the reliance on a given student, we could assume that the student $S$
is drawn from a distribution of students $\Pr(S)$, 
and extend our framework by considering 
the expected benefit for a random student 
averaged over various values of $n$.
In practice, we experiment 
with a small set of diverse students 
(e.g., models with different sizes, architectures, learning procedures) 
and consider different values of $n$.

\subsection{Automated Teachers and Students}
\label{subsec:automated_teachers}

In principle, $T$ and $S$ could be 
either people or algorithms. 
However, 
quantitative measurements are easier to conduct 
when $T$ and (especially) $S$ are algorithms.  
In particular, imagine that $T$ (which for example could be a BERT-based classifier) 
identifies  
an explanation $\bm{e}_{T}(\bm{x})$ 
that is some subset of tokens in a document $\bm{x}$ 
that are relevant to the prediction 
(acquired by, for example,
any of the explanation methods
mentioned in the introduction) 
and $S$ is some machine learner 
that makes use of the explanation.  
The value of teacher-explanations for $S$ 
can then be assessed via standard evaluation 
of explanation-aware student learners, 
using predicted labels instead of gold labels.  
This value can then be compared to other schemes 
for producing explanations
(e.g.,~integrated gradients). %
Albeit, an important concern in automated evaluation  
is that, by design,
the obtained results 
are contingent 
on the student model(s) and how explanations are incorporated by the student model(s). %

Another apparent ``bug'' in this framework 
is that in the automated case, 
one could obtain a perfect simulation accuracy 
with an explanation that communicates 
all the weights of the teacher classifier $f_T$ to the student.\footnote{All 
the weights of the model can be thought of as a complete explanation, 
and is a reasonable choice for simpler models, 
e.g.,~ a linear-model with a few parameters.}  
We propose two approaches 
to address this problem.  
First, we simply limit explanations 
to be of a form that people can comprehend---e.g., 
spans in a document $\bm{x}$.
That is, we 
consider only popular formats of explanations that are 
considered to be human understandable 
(see~\S\ref{sec:learning_with_explanations} for details and Table~\ref{tbl:explanation_example} for examples).
Secondly, we experiment with
a diverse set of student models 
(e.g.,~networks with architectures different
from the original teacher model) 
which precludes trivial weight-copying solutions.

\begin{table*}[ht]
    \small
    \centering
    \begin{tabular}{@{}c@{}}
    \toprule
        \textbf{Sentiment Analysis}~\cite{zaidan2007using}                \\ \toprule
        \begin{tabular}[l]{@{}l@{}} I don't know what movie the critics saw, but it wasn't this one. The popular consensus among newspaper critics \\ was that this movie is unfunny and dreadfully boring. In my personal opinion, \hlnew{\textbf{they couldn't be more wrong}}\\ If you were expecting Airplane! - like laughs and Agatha Christie - intense mystery, then yes, this movie would \\ be a disappointment. However, if you're just looking for an enjoyable movie and a good time, \hlnew{\textbf{this is one to see}} \end{tabular} \\ \midrule
        \textbf{Question Answering}~\cite{lamm2020qed}                                                                                             \\ \midrule
        \begin{tabular}[l]{@{}l@{}} \textbf{Question}: who plays \hlp{\textbf{mabel}}'s voice on \hln{\textbf{gravity falls}}\\
            \textbf{Passage}: Kristen Joy Schaal (born ...) is an American actress, voice actress, comedian and writer. She is best\\known for her roles of Mel in Flight of the Conchords, the over-sexed nurse Hurshe Heartshe on The Heart, She \\ Holler, Louise Belcher in Bob 's Burgers, \hlp{\textbf{Mabel Pines}} in \hln{\textbf{Gravity Falls}}, and Carol in The Last Man on Earth.
        \end{tabular} \\ \midrule
    \end{tabular}
    \caption{Example of annotated rationales in sentiment analysis and referential equalities in QA.
    }
    \label{tbl:explanation_example}
\end{table*}

\subsection{Discussion}
\label{subsec:discussion}

In our framework, two design choices are crucial: 
(i) students do not have access to explanations at test time; 
and (ii) we use a machine learning model 
as a substitute for student learner. 
These two design choices differentiate our framework 
from similar communication games proposed 
by \citet{treviso2020towards} 
and \citet{hase-bansal-2020-evaluating}. 
When explanations are available at test time, 
they can \emph{leak} the teacher output 
directly or indirectly, 
thus corrupting the simulation task. 
Both genuine and trivial explanations 
can encode the teacher output, 
making it difficult to discern 
the quality of explanations.\footnote{A 
trivial explanation may highlight 
the first input token if the teacher output is $0$, 
and the second token if the output is $1$. 
Such explanations, termed as ``Trojan explanations'', 
are a problematic manifestation of several approaches,
as discussed in ~\cite{chang2020invariant, jacovi2020aligning}.} 
The framework of \citet{treviso2020towards}
is affected by this issue, 
which is probably only partially addressed by 
enforcing constraints on the student. 
Preventing access to explanations while testing  
solves this problem and offers flexibility 
in choosing student models.

Substituting machine learners for people allows us 
to train student models on thousands of examples, 
in contrast to the study by~\citet{hase-bansal-2020-evaluating}, 
where (human) students were trained 
on only $16$ or $32$ examples.  
As a consequence, the observed differences 
among many explanation techniques
were statistically insignificant in their studies.
While human subject experiments are 
a valuable complement to scalable automatic evaluations, 
it is expensive to conduct sufficiently large-scale studies; 
and people's preconceived notions 
might impair their ability 
to simulate the models accurately;\footnote{We 
speculate this effect to be pronounced when the models' outputs 
and the true labels differ only over a few samples.} 
and lastly these preconceived notions
might bias performance for different people differently.

\section{Learning with Explanations}
\label{sec:learning_with_explanations}

Our student-teacher framework does not specify
how to use  explanations
while training the student model.  
Below, we examine two broad approaches 
to incorporate explanations: 
attention regularization and multitask learning.
Our first approach regularizes 
attention values of the student model 
to align with the information communicated in explanations.
In the second method,   
we pose the learning task for the student 
as a joint task of prediction and explanation generation, 
expecting to improve prediction 
due to the benefits of multitask learning.  
We show that both of these   
methods indeed improve student performance
when using human-provided explanations 
(and gold labels) for classification tasks.
We explore variants of these two approaches for question answering tasks.

\paragraph{Classification Tasks} 
The training data for the 
student model consists of $n$ documents 
$\bm{x}_1, \dots, \bm{x}_n$, 
and the output to be learned, $y_1, \dots, y_n$,
comes from the teacher, i.e. $y_i=f_T(\bm{x}_i)$, 
along with teacher-explanations 
$\bm{e}_T(\bm{x}_1), \dots, \bm{e}_T(\bm{x}_n)$.
In this work, we consider teacher explanations in the form of 
a binary vector $\bm{e}_T(\bm{x_i})$,
such that $\bm{e}_T(\bm{x_i})_j = 1$ 
if the $j^\text{th}$ token in document $\bm{x_i}$ 
is a part of the teacher-explanation, and $0$ otherwise (see Table~\ref{tbl:explanation_example} for an example).\footnote{Explanations that generate a continuous ``importance'' score for each token can also be used as per this definition, e.g., by selecting the top-$k\%$ tokens from those scores.}
To incorporate explanations during training, 
we suggest two different approaches.
First, we use \textbf{attention regularization}, 
where we add a regularization term
to our loss to reduce the KL divergence 
between the attention distribution 
of the student model ($\boldsymbol{\alpha}_{\text{student}}$)
and the distribution of the teacher-explanation
($\boldsymbol{\alpha}_{\text{exp}}$):
\begin{equation}
    \label{eqn:attention_regularization}
    \mathcal{R}' = - \lambda~ \text{KL}(\boldsymbol{\alpha}_{\text{exp}} \parallel \boldsymbol{\alpha}_{\text{student}}),
\end{equation}
where the explanation distribution ($\boldsymbol{\alpha}_{\text{exp}}$)
is uniform over all the tokens in the explanation 
and $\epsilon$ elsewhere (where $\epsilon$ is a very small constant). 
When dealing with student models that employ multi-headed attention, 
which use multiple different attention vectors at each layer of the model \cite{vaswani2017attention}, 
we take $\boldsymbol{\alpha}_{\text{student}}$ to be 
the attention from the \texttt{[CLS]} token 
to other tokens in the last layer,
averaged across all attention heads.
Several past approaches have used  
attention regularization 
to incorporate 
human rationales,
with an aim to improve 
the overall performance of the system
for classification tasks~\cite{bao-etal-2018-deriving, zhong2019fine}
and machine translation~\cite{yin21acl}.

Second, we use explanations via \textbf{multitask learning}, 
where the two tasks are prediction 
and explanation generation (a sequence labeling problem). 
Formally, the overall loss can be written as: 
\begin{align*}
L &= - \sum_{i=1}^n\left[\log\underbrace{p(y_i | \bm{x_i}; \theta)\ }_\text{classify} + \  \log \underbrace{p(\bm{e_i} | \bm{x_i}; \phi,\theta)}_\text{explain}\right]
\end{align*}

As in multitask learning, 
if the task of prediction and explanation generation are complementary,
then the two tasks would benefit from each other.
As a corollary, if the teacher-explanations 
offer no additional information about the prediction, 
then we would
see no benefit from multitask learning (appropriately so).
For most of our classification experiments,
we use BERT~\cite{devlin-etal-2019-bert} %
with a linear classifier on top of the \texttt{[CLS]}
vector
to model $p(y|\bm{x}; \theta)$.
To model $p(\bm{e}|\bm{x}; \phi~\theta)$
we use a linear-chain CRF~\cite{lafferty2001conditional}
on top of the sequence vectors from BERT.
Note that we share the BERT parameters $\theta$
between classification and explanation tasks. 
In prior work, similar multitask formulations  
have been demonstrated to effectively incorporate rationales 
to improve classification performance \cite{zaidan-eisner-2008-modeling}
and evidence extraction \cite{pruthi2020weakly}.

\paragraph{Question Answering}
Let the question $q$ consist of m tokens $q_1 \dots q_m$, 
along with passage $\bm{x}$
that provides the answer to the question,
consisting of $n$ tokens $\bm{x}_1, \dots, \bm{x}_n$. 
Let us define a set of question phrases $\mathcal{Q}$ 
and passage phrases $\mathcal{P}$ to be 
\begin{eqnarray*}
   \mathcal{Q} &= \{(i, j):1 \leq i \leq j \leq m\} \\
   \mathcal{P} &= \{(i, j):1 \leq i \leq j \leq n\}.
\end{eqnarray*}
We consider a subset of QED explanations \cite{lamm2020qed},
which consist of a sequence
of one or more ``referential equality annotations'' $e_1 \dots e_{|e|}$.
Formally, each referential equality annotation $e_k$ for $k=1 \ldots |e|$ is a pair $(\phi_k, \pi_k) \in \mathcal{Q} \times \mathcal{P}$, 
specifying that phrase $\phi_k$ in the question 
refers to the same thing in the world 
as the phrase $\pi_k$ in the passage 
(see Table~\ref{tbl:explanation_example} for an example).

To incorporate explanations for question answering tasks, 
we use the two approaches discussed 
for text classification tasks,
namely attention regularization and multitask learning.
Since the explanation format for question answering   
is different from the explanations in text classification,
we use a lossy transformation, where 
we construct a binary explanation vector, 
where $1$ corresponds to tokens that appear 
in one or more referential equalities and $0$ otherwise.
Given the transformation, 
both these approaches do not use 
the alignment information present
in the referential equalities.

To exploit the alignment information 
provided by referential equalities, 
we introduce and append the standard loss with
\textbf{attention alignment loss}:
\begin{equation*}
    \mathcal{R}^{'} = -\lambda \log\left(\frac{1}{|e|} \sum_{k=1}^{|e|}\boldsymbol{\alpha}_{\text{student}}[\phi_k \rightarrow \pi_{k}]\right), 
\end{equation*}
where  $e_k = (\phi_k, \pi_k)$ is the $k^\text{th}$ referential equality,  
and $\boldsymbol{\alpha}_{\text{student}}[\phi_k \rightarrow \pi_k])$ 
is the last layer average attention originating 
from tokens in $\phi_k$ to tokens in $\pi_k$.
The average is computed across all the tokens in $\phi_k$ 
and across all attention heads. 
The underlying idea is to increase attention values 
corresponding to the alignments provided in explanations.

\section{Human Experts as Teachers}
\label{sec:results}

Below, we discuss the results 
upon applying our framework
to explanations and output from human teachers 
to confirm if expert explanations 
improve the student models' performance.

\paragraph{Setup} 
There exist a few tasks where researchers 
have collected explanations 
from experts besides the output label.
For the task of sentiment analysis on movie reviews, %
~\citet{zaidan2007using}
collected ``rationales''
where people highlighted portions of the 
movie reviews that would encourage (or discourage) 
readers to watch (or avoid) the movie.
In another recent effort, 
\citet{lamm2020qed} collected ``QED annotations'' 
over questions and the passages 
from the Natural Questions (NQ)
dataset~\cite{kwiatkowski2019natural}. 
These annotations contain 
the salient entity in the question
and their referential mentions
in the passage that need to be resolved 
to answer the question. 
For both these tasks, our student-learners 
are pretrained BERT-base models, 
which are further fine-tuned 
with outputs and explanations from human experts.

\begin{table}[t]
\centering
\begin{tabular}{@{}lrrr@{}}
\toprule
Student Model & 600 & 900 & 1200 \\ \midrule 
BERT-base & 75.5 & 79.0 &  81.1 \\  \hline
w/ explanations used via & & & \\  
\:  multitask learning & 75.2 & 80.0  & 82.5 \\  
\:  attention regularization & 81.5 & 83.1 & 84.0 \\  \bottomrule

\end{tabular}
    \caption{Simulation accuracy of a student model
    when trained with and without explanations
    from human experts for \textbf{sentiment analysis}.
    We note that both the proposed methods
    to learn with explanation improve performance:
    attention regularization leads to large gains, 
    whereas multitask learning 
    requires more examples to yield improvements.}
    \label{tbl:people_classification}
\end{table}

\paragraph{Results} Our suggested 
methods to learn from explanations 
indeed benefit from human explanations.
For the sentiment analysis task, 
attention regularization 
boosts performance, as depicted 
in Table~\ref{tbl:people_classification}.
For instance, attention regularization improves the accuracy 
by an absolute $6$ points, for $600$ examples. 
The performance benefits, unsurprisingly, 
diminish with increasing training examples---for 
$1200$ examples, the attention regularization 
improves performance by $2.9$ points.
While attention regularization is immediately effective, 
the multitask learning 
requires more examples 
to learn the sequence labeling task.
We do not see any improvement 
using multitask learning for $600$ examples,
but for $900$ and $1200$ training examples,
we see absolute improvements of $1$ and $1.4$ points, respectively.

We follow up our findings to validate 
if the simulation performance of the student model
is correlated with explanation quality.
To do so, we corrupt human explanations 
by unselecting the marked tokens 
with varying noising probabilities 
(ranging from 0 to 1, in steps of 0.1).
We train student models on corrupted explanations 
using attention regularization 
and find their performance 
to be highly negatively correlated 
with the amount of noise
(Pearson correlation $\rho=-0.72$).  
This study verifies that 
our metric is correlated with 
(an admittedly simple notion of) explanation quality.

For the question-answering task, 
we measure the F1 score
of the student model
on the test set carved from the QED dataset.
As one can observe from Table~\ref{tbl:people_qa}, %
both attention regularization and 
attention alignment loss
improve the performance, 
whereas multitask learning is not effective.\footnote{We 
speculate that multitask learning might require more than $2500$ examples to yield benefits. Unfortunately, for the QED dataset, we only have $2500$ training examples.}
Attention regularization and attention alignment loss 
improve F1 score by $2.3$ and $8.4$ points 
for $500$ examples, respectively. 
The gains decrease with increasing examples 
(e.g., the improvement due to attention alignment loss 
is $5$ points on $2500$ examples, 
compared to $8.4$ points with $500$ examples).
The key takeaway from these experiments 
(with explanations and outputs from human experts) 
is that we observe benefits 
with the learning procedures  
discussed in previous section. 
This provides support to our proposal 
to use these methods %
for evaluating various explanation techniques.

\begin{table}[t!]
\centering
\begin{tabular}{@{}lrrr@{}}
\toprule
Student Model & 500 & 1500 & 2500 \\ \midrule 
BERT-base & 28.9 & 43.7 &  49.0 \\  \hline
w/ explanations used via & & & \\  
\:  multitask learning & 29.7 & 42.7  & 49.2 \\  
\:  attention regularization & 31.2 & 47.2 & 52.6 \\  
\:  attention alignment loss & 37.3 & 49.6 & 54.0 \\ \bottomrule 

\end{tabular}
    \caption{Simulation performance (F1 score) of a student model 
    when trained with and without explanations 
    from human experts for \textbf{question answering}.
    We find that attention regularization 
    and attention alignment loss 
    result in large improvements 
    upon incorporating explanations.}
    \label{tbl:people_qa}
\end{table}

\section{Automated Evaluation of Attributions}
\label{subsec:results_explanations}

Here, we use a machine learning model as 
our choice for the teacher, 
and subsequently train student models 
using the output and explanations 
produced by the teacher model.
Such a setup allows us 
to compare attributions 
produced by different techniques 
for a given teacher model.

\begin{table*}[t]
    \centering
    \begin{tabular}{lrrr|rrr?rrr|rrr}
    \toprule
        & \multicolumn{6}{c?}{\cellcolor[HTML]{F5F5F5}{\textbf{BERT-base Student}}} & \multicolumn{6}{c}{{\textbf{BERT-large Student}}} \\  \hline
                         & \multicolumn{3}{c|}{\cellcolor[HTML]{F5F5F5}\footnotesize Attn. Regularization} &   \multicolumn{3}{c?}{\cellcolor[HTML]{F5F5F5}\footnotesize Multitask Learning} & \multicolumn{3}{c|}{\footnotesize Attn. Regularization}   & \multicolumn{3}{c}{\footnotesize Multitask Learning} \\
                            \cline{2-7} \cline {8-13}
    Explanations             & \cellcolor[HTML]{F5F5F5}1000    & \cellcolor[HTML]{F5F5F5}2000 & \cellcolor[HTML]{F5F5F5}\footnotesize{Rank}   & \cellcolor[HTML]{F5F5F5}4000 & \cellcolor[HTML]{F5F5F5}8000 & \cellcolor[HTML]{F5F5F5}\footnotesize{Rank}   & 1000 & 2000 & \footnotesize{Rank} & 4000 & 8000 & \footnotesize{Rank} \\ \midrule
    None       & \cellcolor[HTML]{F5F5F5}91.5    & \cellcolor[HTML]{F5F5F5}92.6   & \cellcolor[HTML]{F5F5F5}6.0  &   \cellcolor[HTML]{F5F5F5}93.6 & \cellcolor[HTML]{F5F5F5}94.9 & \cellcolor[HTML]{F5F5F5}6.0  & 92.6      & 93.0 & 8.5 & 93.7 & 94.5 & 8.0  \\
    Random    &  \cellcolor[HTML]{F5F5F5} \underline{90.6}    & \cellcolor[HTML]{F5F5F5} 92.4 & \cellcolor[HTML]{F5F5F5}8.5 &  \cellcolor[HTML]{F5F5F5}94.1 & \cellcolor[HTML]{F5F5F5}\underline{94.5} & \cellcolor[HTML]{F5F5F5}5.5  & 92.4 & 93.1 & 7.5 & 94.3 & 94.4 & 7.5 \\ 
    Trivial   & \cellcolor[HTML]{F5F5F5} \underline{82.8}    &  \cellcolor[HTML]{F5F5F5}\underline{88.3} & \cellcolor[HTML]{F5F5F5}10.0  & \cellcolor[HTML]{F5F5F5} \underline{93.4} & \cellcolor[HTML]{F5F5F5} \underline{93.8} & \cellcolor[HTML]{F5F5F5}9.5 & 90.3 & 92.7 & 10.0 & 93.1 & 94.7 & 8.5 \\ \hline
    LIME                 &  \cellcolor[HTML]{F5F5F5}91.3    & \cellcolor[HTML]{F5F5F5}92.6   & \cellcolor[HTML]{F5F5F5} 7.0 & \cellcolor[HTML]{F5F5F5}94.0    & \cellcolor[HTML]{F5F5F5}95.0 & \cellcolor[HTML]{F5F5F5} 5.0 & \underline{92.7} & 93.1 & 7.5 & \underline{94.6} & \underline{95.1} & 4.0   \\
    Grad. Norm        &  \cellcolor[HTML]{F5F5F5}91.6    & \cellcolor[HTML]{F5F5F5}92.4   & \cellcolor[HTML]{F5F5F5}6.5 & \cellcolor[HTML]{F5F5F5}\underline{94.3} & \cellcolor[HTML]{F5F5F5}94.2 & \cellcolor[HTML]{F5F5F5}6.0 & \underline{92.9} & 93.1 & 6.5 & 93.8 & 94.0 & 8.5 \\
    Grad$\times$Inp    & \cellcolor[HTML]{F5F5F5}91.7    & \cellcolor[HTML]{F5F5F5}92.2   & \cellcolor[HTML]{F5F5F5}7.0 & \cellcolor[HTML]{F5F5F5}\underline{94.4} & \cellcolor[HTML]{F5F5F5}94.5 & \cellcolor[HTML]{F5F5F5}5.0 & 93.0 & \underline{93.7} & 5.0 & 93.5 & \underline{94.8} & 7.0 \\
    DeepLIFT      & \cellcolor[HTML]{F5F5F5}\underline{92.0} & \cellcolor[HTML]{F5F5F5}\underline{93.4} & \cellcolor[HTML]{F5F5F5}4.0 & \cellcolor[HTML]{F5F5F5}92.0 & \cellcolor[HTML]{F5F5F5}94.1 & \cellcolor[HTML]{F5F5F5}9.5 & \underline{93.6} & \underline{94.2} & 2.5 & \underline{94.8} & \underline{95.3} & 2.0 \\
    Layer Cond.      & \cellcolor[HTML]{F5F5F5}92.3 & \cellcolor[HTML]{F5F5F5}\underline{93.5} & \cellcolor[HTML]{F5F5F5}3.0 & \cellcolor[HTML]{F5F5F5}\underline{94.1} &\cellcolor[HTML]{F5F5F5}94.7 & \cellcolor[HTML]{F5F5F5}5.5 & \underline{93.4} & \underline{94.1} & 4.0 & \underline{94.6} & 94.7 & 4.5 \\
    I.G. & \cellcolor[HTML]{F5F5F5}\underline{92.6}    & \cellcolor[HTML]{F5F5F5}\underline{93.6}   & \cellcolor[HTML]{F5F5F5}2.0  & \cellcolor[HTML]{F5F5F5}\underline{94.5} & \cellcolor[HTML]{F5F5F5}95.2 & \cellcolor[HTML]{F5F5F5}2.0  & \underline{93.4} & \underline{94.4} & 2.5 & \underline{94.6} & \underline{95.1} & 4.0 \\
    Attention &  \cellcolor[HTML]{F5F5F5}\underline{\textbf{93.9}} & \cellcolor[HTML]{F5F5F5}\underline{\textbf{95.2}} &\cellcolor[HTML]{F5F5F5}\textbf{1.0}  & \cellcolor[HTML]{F5F5F5}\underline{\textbf{96.0}} & \cellcolor[HTML]{F5F5F5}\underline{\textbf{96.6}} & \cellcolor[HTML]{F5F5F5}\textbf{1.0} & \underline{\textbf{94.0}} & \underline{\textbf{95.4}} & \textbf{1.0} & \underline{\textbf{95.4}} & \underline{\textbf{96.1}} & \textbf{1.0}  \\
    \bottomrule
    \end{tabular}
    \caption{We evaluate the effectiveness of attribution methods for sentiment analysis using simulation accuracy of student models trained with these explanations on varying amounts of data (\S\ref{subsec:student_teacher_results}). Each method selects top-10\% ``important'' tokens for each example. We find attention-based explanations to be most effective, 
    followed by integrated gradients. We also tabulate the average rank as per our metric. Statistically significant differences (p-value $< 0.05$) from the no-explanation control are \underline{underlined}. }
    \label{tbl:results_movie_reviews}
\end{table*}

\subsection{Setup}
For sentiment analysis, 
we use BERT-base~\cite{devlin-etal-2019-bert} as our teacher model 
and train it on the IMDb dataset~\cite{maas-EtAl:2011:ACL-HLT2011}. 
The accuracy of the teacher model is 93.5\%.
We compare seven 
commonly used methods for producing explanations. 
These techniques
include LIME~\cite{ribeiro2016should}, 
gradient-based saliency methods, 
i.e.,~gradient norm and gradient $\times$ input \cite{simonyan2013deep}, 
DeepLIFT \cite{shrikumar2017learning}, 
layer conductance \cite{dhamdhere2018important},
integrated gradients (I.G.)  \cite{sundararajan2017axiomatic},
and attention-based explanations \cite{bahdanau2014neural}. More details 
about these explanation techniques are provided in the Appendix. 

For each explanation technique 
to be comparable to others, 
we sort the tokens as per scores assigned 
by a given explanation technique, 
and use only the top-$k\%$ tokens.
This also ensures that across different explanations, 
the quantity of information from the teacher 
to the student per example is constant.
Additionally, we evaluate no-explanation,  
random-explanation and trivial-explanation baselines. 
For random explanations, we randomly choose 
$k\%$ tokens, and for trivial explanations,
we use the first $k\%$ tokens for the positive class,
and the next $k\%$ tokens for the negative class. 
Such trivial explanations encode the label 
and can achieve perfect scores 
for many evaluation protocols 
that use explanations at test time. 

Corresponding to each explanation type, 
we train 4 different student models---comprising 
BERT and BiLSTM based models---using
outputs and explanations from the teacher. 
The test set of the teacher model 
is divided to construct train, development,
and test splits for the student model.
We train student models with explanations 
by using attention regularization and multitask learning. 
We vary the amount of training data available 
and note the simulation performance of student models.

For the question answering task, 
we use the Natural Questions dataset~\cite{kwiatkowski2019natural}.
The teacher model is a SpanBERT-based model 
that is trained jointly to answer the question 
and produce explanations~\cite{lamm2020qed}. 
We use the model made available by the authors.
The test set of Natural Questions is split to form 
the training, development and test set for the student model. 
We use a BERT-base QA model as our student model
to evaluate the value of teacher explanations.

\begin{table*}[ht]
\centering
\begin{tabular}{@{}lrrr?rrr|rrr@{}}
\toprule
& \multicolumn{3}{c?}{\cellcolor[HTML]{F5F5F5}{\textbf{Bi-LSTM Student}}} & \multicolumn{6}{c}{{\textbf{Bi-LSTM+Attention Student}}} \\  \hline
    & \multicolumn{3}{c?}{\cellcolor[HTML]{F5F5F5}Multitask Learning} & \multicolumn{3}{c|}{{Attn. Regularization}} & \multicolumn{3}{c}{{Multitask Learning}}    \\ 
\cline{2-4} \cline{5-10} 
    Explanations &  \cellcolor[HTML]{F5F5F5}4000 & \cellcolor[HTML]{F5F5F5}8000     &  \cellcolor[HTML]{F5F5F5}Rank  & 1000 & 2000 &  Rank &  4000 & 8000 & {Rank}\\ \midrule
No Explanation       & \cellcolor[HTML]{F5F5F5}71.1      & \cellcolor[HTML]{F5F5F5}85.1    &  \cellcolor[HTML]{F5F5F5}10.0 & 78.2 & 82.1 & 9.0 & 85.8 & 88.5 & 9.0   \\
Random       & \cellcolor[HTML]{F5F5F5}75.8     & \cellcolor[HTML]{F5F5F5}86.4   &  \cellcolor[HTML]{F5F5F5}7.0 & 78.2  & 82.2  & 8.0  & 86.0  & 88.5  & 8.0    \\
Trivial       & \cellcolor[HTML]{F5F5F5}77.4      & \cellcolor[HTML]{F5F5F5}85.7    &  \cellcolor[HTML]{F5F5F5}6.5 & \underline{68.0}  & \underline{72.4}  & 10  & 85.5  & 88.2  & 10.0   \\ \hline
LIME & \cellcolor[HTML]{F5F5F5}\underline{77.1}      & \cellcolor[HTML]{F5F5F5}\underline{87.1}    &  \cellcolor[HTML]{F5F5F5}5.0 & \underline{79.5} & \underline{83.7} & 5.5 & 86.2 & \underline{89.1} & 6.0   \\
Gradient Norm & \cellcolor[HTML]{F5F5F5}\underline{77.3}      & \cellcolor[HTML]{F5F5F5}\underline{86.9}    &  \cellcolor[HTML]{F5F5F5}5.0 & 79.0 & \underline{83.9} & 5.0 & \underline{87.1} & \underline{89.2} & 4.0   \\
Gradient $\times$ Input & \cellcolor[HTML]{F5F5F5}\underline{77.0}      & \cellcolor[HTML]{F5F5F5}85.8    &  \cellcolor[HTML]{F5F5F5}7.5 & 78.7 & \underline{83.8} & 6.0 & 86.2 & 88.8 & 6.5   \\
DeepLIFT       & \cellcolor[HTML]{F5F5F5}\underline{74.8}     & \cellcolor[HTML]{F5F5F5}\underline{86.1}    &  \cellcolor[HTML]{F5F5F5}8.0 & \underline{80.2} & \underline{83.5} & 5.5 & \underline{86.6} & \underline{89.1} & 5.5   \\
Layer Conductance       & \cellcolor[HTML]{F5F5F5}\underline{\textbf{79.5}}     & \cellcolor[HTML]{F5F5F5}\underline{87.4}    &  \cellcolor[HTML]{F5F5F5}\textbf{1.5} & \underline{\textbf{81.7}} & \textbf{\underline{85.5}} & \textbf{1.5} & \underline{87.3} & \underline{89.4} & 2.5   \\
Integrated Gradients       & \cellcolor[HTML]{F5F5F5}\underline{78.8}      & \cellcolor[HTML]{F5F5F5}\underline{87.3}    & \cellcolor[HTML]{F5F5F5}2.5 & \underline{80.8} & \textbf{\underline{85.5}} & 2.0 & \underline{87.3} & \underline{89.3} & 2.5   \\
Attention       & \cellcolor[HTML]{F5F5F5}\underline{78.1}      & \cellcolor[HTML]{F5F5F5}\underline{\textbf{88.3}}   & \cellcolor[HTML]{F5F5F5}2.0 & \underline{81.1} & \underline{84.6} & 2.5  & \underline{\textbf{87.4}} & \underline{\textbf{89.8}} & \textbf{1.0}   \\ \bottomrule
\end{tabular}
    \caption{Evaluating different attribution methods for sentiment analysis 
    using the simulation accuracy of BiLSTM-based student models 
    trained with these explanations on varying amounts of data (\S\ref{subsec:student_teacher_results}). 
    We find attention-based explanations, integrated gradients,
    and layer conductance to be effective techniques.
    The rankings are largely consistent 
    with those attained using transformer-based student models (Table~\ref{tbl:results_movie_reviews}). 
    Statistically significant differences (p-value $< 0.05$)
    from the no-explanation control are \underline{underlined}. }
    \label{tbl:bilstm_students}
\end{table*}

\subsection{Main Results}
\label{subsec:student_teacher_results}
We evaluate different explanation generation methods 
based upon the simulation accuracy of various student models
for two NLP tasks: text classification and question answering.

For the \textbf{sentiment analysis task}, 
we present the simulation performance 
of BERT-base and BERT-large student models 
in Table~\ref{tbl:results_movie_reviews}, 
and BiLSTM and BiLSTM+Attention student models 
in Table~\ref{tbl:bilstm_students}.
From these two tables, we first note that
attention-based explanations are effective, 
resulting in large and statistically significant 
improvements over the no-explanation control. 
We see an improvement of 1.4 to 2.6 points 
for transformer-based student models 
(Table~\ref{tbl:results_movie_reviews}), 
and up to 7 points for the Bi-LSTM student model
(Table~\ref{tbl:bilstm_students}).

While it may seem that attention is effective 
because it aligns most directly 
with attention regularization learning strategy,
we note that the trends from multitask learning 
corroborate the same conclusion 
for different student models---especially 
the Bi-LSTM student model, which does not
even use the attention mechanism, 
and therefore can not incorporate explanations 
using attention regularization.
Besides attention explanations,
we also find integrated gradients 
and layer conductance to be effective techniques. 
Qualitatively inspecting a few examples, 
we notice that attention and integrated gradients 
indeed highlight spans that convey the sentiment of the review.

Lastly, we see that trivial explanations do not
outperform the control experiment, 
confirming that our framework is robust 
to such gamification attempts. 
These explanations would result in a perfect score 
for the protocol discussed in~\cite{treviso2020towards}. 
The metric by ~\citet{hase2020leakage} 
would be undefined in the case 
when 100\% of the explanations 
trivially leak the label---in the limiting case 
(when all but one explanation leak the label trivially), 
the metric would result in a high score, which is unintended.

For the \textbf{question answering task}, %
we observe from Table~\ref{tbl:teacher_qa}
that explanations from SpanBERT QA model are effective, 
as indicated by both the approaches 
to learn from explanations. %
The performance benefit using attention alignment loss 
with $250$ examples is $7.9$ absolute points, 
and these gains decrease (unsurprisingly)
with increasing number of training examples. 
For instance, the gain is only $2.9$ points for $4000$ examples and the benefits vanish with $16000$ training examples. 

\begin{table}
\centering
\begin{tabular}{@{}lrrrr@{}}
\toprule
Student Model & 250 & 1K & 4K & 16K \\ \midrule 
BERT & 25.0 & 37.7 &  52.6 & 61.6 \\  \hline
\footnotesize w/ explanations used via & \\ 
\:  \footnotesize attention regularization & 27.6 & 42.8 & 54.4 & 62.3\\    
\:  \footnotesize attention alignment loss & 32.9 & 46.9 & 55.5 & 62.2 \\ \bottomrule 
\end{tabular}
    \caption{Simulation performance (F1 score) of a student model when trained with and without explanations from the SpanBERT QA model (the teacher model in this case). We find these explanations to be effective across both the learning strategies.}
    \label{tbl:teacher_qa}
\end{table}

\begin{table*}[t]
  \centering
  \begin{tabular}{l|rrr|rrr|rrr|rrr}
  \toprule
   &
    \multicolumn{10}{c}{\textbf{Bi-LSTM w/ Attention Student Models}} \\ \hline
  \multicolumn{1}{c|}{\begin{tabular}[c]{@{}c@{}}{\small BS, ED, HS}\\{\small LR} \end{tabular}} &
    \multicolumn{3}{c|}{\begin{tabular}[c]{@{}c@{}}16, 128, 256,\\ $2.5\times10^{-3}$ \end{tabular}} &
    \multicolumn{3}{c|}{\begin{tabular}[c]{@{}c@{}}16, 64, 256,\\ $0.5\times10^{-2}$ \end{tabular}} &
    \multicolumn{3}{c|}{\begin{tabular}[c]{@{}c@{}}64, 256, 256,\\ $0.5\times10^{-2}$ \end{tabular}} &
    \multicolumn{3}{c}{\begin{tabular}[c]{@{}c@{}}32, 128, 768,\\ $2.5\times10^{-3}$ \end{tabular}} \\ \hline
    &
    {\footnotesize MTL} &
    {\footnotesize AR} &
    {\footnotesize Rank} &
    {\footnotesize MTL} &
    {\footnotesize AR} &
    {\footnotesize Rank} &
    {\footnotesize MTL} &
    {\footnotesize AR} &
    {\footnotesize Rank} &
    {\footnotesize MTL} &
    {\footnotesize AR} &
    {\footnotesize Rank} \\ \hline
  None &
    83.9 &
    79.7 &
    8.0 &
    84.1 &
    79.9 &
    7.5 &
    84.6 &
    80.0 &
    8.0 &
    82.3 &
    77.4 &
    8.0 \\
  LIME &
    {\ul 84.3} &
    80.3 &
    5.0 &
    {\ul 84.9} &
    80.8 &
    5.0 &
    85.0 &
    {\ul 81.0} &
    4.5 &
    {\ul 83.4} &
    {\ul 79.4} &
    5.5 \\
  Grad Norm &
    84.1 &
    80.5 &
    5.0 &
    {\ul 85.0} &
    {\ul 81.2} &
    4.0 &
    84.7 &
    {\ul 81.5} &
    5.5 &
    83.9 &
    79.4 &
    3.0 \\
  Grad$\times$Inp. &
    84.0 &
    80.0 &
    6.5 &
    {\ul 83.9} &
    80.6 &
    7.5 &
    84.8 &
    {\ul 81.3} &
    5.0 &
    83.4 &
    79.0 &
    6.5 \\
  DeepLIFT &
    84.0 &
    {\ul 81.2} &
    5.5 &
    {\ul 84.6} &
    81.7 &
    5.0 &
    84.9 &
    {\ul 82.1} &
    3.5 &
    83.5 &
    {\ul 79.0} &
    5.5 \\
  Layer Cond. &
    {\ul 84.7} &
    {\ul 82.0} &
    3.0 &
    {\ul 85.1} &
    {\ul \textbf{83.8}} &
    \textbf{1.5} &
    84.7 &
    {\ul 82.3} &
    4.5 &
    {\ul 83.7} &
    {\ul 80.2} &
    3.5 \\
  I.G. &
    {\ul \textbf{84.8}} &
    {\ul 82.3} &
    \textbf{1.5} &
    {\ul 84.9} &
    {\ul 83.6} &
    3.5 &
    84.8 &
    {\ul \textbf{82.4}} &
    3.0 &
    {\ul \textbf{84.1}} &
    {\ul 80.3} &
    \textbf{1.5} \\
  Attention &
    {\ul 84.7} &
    {\ul \textbf{82.6}} &
    \textbf{1.5} &
    {\ul \textbf{85.5}} &
    {\ul 81.8} &
    2.0 &
    \textbf{85.3} &
    {\ul 82.1} &
    \textbf{2.5} &
    {\ul 83.7} &
    {\ul \textbf{80.6}} &
    2.0 \\
    \bottomrule
  \end{tabular}
  \caption{Gauging the sensitivity of our framework to different hyperparameter values of student models. We note the simulation accuracies of $4$ BiLSTM with attention student models with varying batch size (\textbf{BS}), learning rate (\textbf{LR}), embedding dimension (\textbf{ED}), and hidden size (\textbf{HS}). We incorporate explanations via multi-task learning (MTL) over $4$K examples and attention regularisation (AR) on $2$K examples. 
  The average rank correlation coefficient $\tau$ between all five configurations (including one from Table \ref{tbl:bilstm_students}) is 0.95.
  Statistically significant differences (p-value $< 0.05$) from the no-explanation control are \underline{underlined}.
  }
  \label{tbl:results_hyperparameter_sensitivity}
  \end{table*}

\subsection{Analysis}

Here, we analyze the 
the effect of different instantiations of our framework, i.e., 
sensitivity to the choice of student architectures, their hyperparameters, learning strategies, etc.
Additionally,
we examine 
the effect of varying the percentage of explanatory tokens ($k$ in top-$k$ tokens)
on the results obtained from our framework.

\paragraph{Varying student models and learning strategies}
We evaluate the agreement among attribution rankings obtained using 
(i) different learning strategies; 
and (ii) different student models.
We compute the Kendall rank correlation coefficient $\tau$ 
to measure the agreement among different attribution rankings.%
\footnote{Note that $\tau$ lies in $[-1, 1]$ with $1$ signifying perfect agreement 
and $-1$ perfect disagreement.} 
We report  
different $\tau$ values 
for varying combinations of student models and learning strategies 
in the Appendix
(Table~\ref{tbl:kendall_tau_coefficient}).
The key takeaways from this investigation are twofold: 
first the rank correlation between rankings 
produced using the two learning strategies---attention regularization (AR) and  multi-task learning (MTL)---for the same student model %
is 0.64 which is considered a high agreement. 
This value is obtained by averaging $\tau$ values from $3$
different student models that can use 
both these learning strategies. 
Second, the rank correlation among rankings 
produced using %
different student models (given the the same learning strategy) is also high---we report average values of 0.65 and 0.47 
when we use 
AR and MTL learning strategies respectively. 
For completion, we also compute $\tau$ for 
all distinct combinations
across student models and learning strategies 
($21$ combinations in total)
and obtain an average value of 0.52.
Overall, we observe high agreement 
among different rankings attained 
through different instantiations 
of our student-teacher framework.

\paragraph{Sensitivity to hyperparameters}
We examine the sensitivity of our framework 
to different hyperparameter values of the student models. %
For BiLSTM-based student models, we
perform a random search over 
different values of four hyperparameters, i.e., 
number of embedding dimensions ($\text{ED} \in \{64, 128, 256, 512\}$), 
number of hidden size ($\text{HS} \in \{256, 512, 768, 1024\}$), 
batch size ($\text{BS} \in \{8, 16, 32, 64\}$) 
and learning rate ($\text{LR} \in \{0.5 \times10^{-3}, 1 \times 10^{-3}, 2.5 \times 10^{-3}, 0.5 \times 10^{-2}\}$).
From all possible configurations above, 
we randomly sample $4$ configurations 
and train a BiLSTM with attention student model 
corresponding to each configuration.
The simulation 
accuracy of student models with different choices of hyperparameters
are presented in Table~\ref{tbl:results_hyperparameter_sensitivity}. 
For a given hyperparameter configuration, we average the ranks across the two learning strategies.
We compute the 
Kendall rank correlation 
coefficient $\tau$ 
amongst rankings 
obtained using different hyperparameter
configurations (including the default configuration 
from Table \ref{tbl:bilstm_students}, thus resulting in $5 \choose 2$ comparisons). 
We obtain a high average correlation of $0.95$ suggesting 
that our framework 
yields largely consistent ranking of attributions %
across varying hyperparameters.

\paragraph{Varying the percentage of explanatory tokens} 
To examine the effect of $k$ in selecting top-$k\%$ tokens, 
we evaluate the simulation performance of BERT-base students 
trained with varying values of $k\in\{5, 10, 20, 40\}$ on $2000$ examples.\footnote{Note that $k$ is not a parameter of our framework, but controls the number of explanatory tokens for each attribution.} 
For these values of $k$, we corroborate the same trend, 
i.e., attention-based explanations are the most effective, 
followed by integrated gradients 
(see Table~\ref{tbl:results_movie_reviews_varying_k} in the Appendix). 
 We also perform an experiment where we consider 
 the entire attention vector to be an explanation, 
 as it does not lose any information due to thresholding. 
 For $500$ examples, we see a statistically significant improvement 
 of $0.9$ over the top-$10$\% attention variant (p-value = $0.03$), 
 the difference shrinks with increasing numbers
 of training examples ($0.1$ for $2000$ examples).

\subsection{Comparison With Other Benchmarks}
For completeness, we compare the ranking 
of explanations obtained through our metrics
with existing metrics of sufficiency and comprehensiveness 
introduced in \cite{deyoung-etal-2020-eraser}. 
The sufficiency metric computes the average difference in the model output 
upon using the input example versus using the explanation alone ($f_T(x) - f_T(e)$), 
while comprehensiveness metric is the average 
of $f_T(x) - f_T(x \backslash e)$ over the examples. 
Note that using these metrics is not ideal 
as they rely upon the model output on deformed input instances 
that lie outside the support of the training distribution.%

We present these metrics for different explanations in Table \ref{tbl:sufficiency_and_comprehensiveness}.  
We observe that LIME outperforms other explanations 
on both the sufficiency and comprehensiveness metrics. 
We attribute this to the fact that LIME explanations 
rely on attributions 
from a surrogate linear model 
trained on perturbed sentences, 
akin to the inputs used to compute these metrics.
The average rank correlation of rankings obtained by our metrics 
(across all students and tasks) 
with the rankings from these two metrics is moderate ($\tau = 0.39$),
which 
indicates that the two proposals produce
slightly different orderings. 
This is unsurprising as  
our protocol, in principle, is different 
from the compared metrics.

Ideally, 
we would like to link this comparison with 
some notion of user preference. 
This aspiration to evaluate %
inferred associations with users is similar to that of evaluating latent topics for topic models~\cite{chang2009leaves}. 
However, directly asking users 
for their preference (for one explanation versus the other)
would be inadequate, as users 
would not be able to comment upon the \emph{faithfulness} 
of the explanation to the computation that resulted in the prediction.  
Instead, we conduct 
a study inspired  
from our protocol, i.e., where users simulate the 
model with and without explanations.

\begin{table}[t]
  \small
  \centering
  \begin{tabular}{@{}lrrrrr@{}}
  \toprule
  & \multicolumn{2}{c}{Sufficiency $\downarrow$} &  & \multicolumn{2}{c}{\footnotesize{Comprehensive.} $\uparrow$} \\  
  \cline{2-3}  \cline{5-6}
      Explanations &  \footnotesize Value &  \footnotesize Rank & & \footnotesize  Value & \footnotesize  Rank \\ \midrule
  Random       &  0.29 & 6 & & 0.04 & 9   \\
  Trivial       &  0.29 & 7 & & 0.04 & 8   \\
  LIME       &  0.06 & 1 & & 0.32 & 1   \\
  Grad Norm &  0.25 & 5 & & 0.11 & 5  \\
  Grad$\times$Inp. & 0.33 & 8 & & 0.06 & 7  \\
  DeepLIFT & 0.39 & 9 & & 0.06 & 6 \\
  Layer Cond. & 0.11 & 2 & & 0.24 & 3   \\
  I.G. & 0.13 & 4 & & 0.17 & 4 \\
  Attention & 0.11 & 3 & & 0.28 & 2  \\ \bottomrule
  \end{tabular}
      \caption{Comparing attribution methods as per the sufficiency (lower the better) and comprehensiveness
      metrics proposed in~\cite{deyoung-etal-2020-eraser}. }
      \label{tbl:sufficiency_and_comprehensiveness}
  \end{table}

\subsection{Human Students}
As  
 discussed in~\S\ref{subsec:discussion}, it is difficult to ``train'' people using a small number of input, output, explanation triples to understand the model sufficiently to simulate the model (on unseen examples) better than the control baseline. A recent study trained students with $16$ or $32$ examples, and tested  if students could simulate the model better using different explanations, however the observed differences among techniques were not statistically significant~\cite{hase-bansal-2020-evaluating}.
Here, we attempt a similar human study, where we present each crowdworker $60$ movie reviews, and for $40$ (out of $60$) reviews we supplement explanations of the model predictions. The goal for the workers is to understand the teacher model and guess the output of the model on the $20$ unseen movie reviews for which explanations are unavailable. 

In our case,  the teacher model accurately predicts 93.5\% of the test examples, therefore to avoid crowdworkers conflating the task of simulation with that of sentiment prediction, we over-sample the error cases  
such that our final setup comprises 50\% correctly classified and 50\% incorrectly classified reviews. 
We experiment with $3$ different attribution techniques: attention (as it is one of the best performing explanation technique as per our protocol), LIME (as it is not very effective according to our metrics, but nonetheless is a popular technique) and random (for control).
We divide a total of $30$ crowdworkers in three cohorts 
corresponding to each explanation type. 
The average simulation accuracy of workers is 68.0\%, 69.0\% and 75.0\% using LIME, attention and random explanations respectively. However, given the large variance in the performance of workers in each cohort, the differences between any pair of these explanations is \textbf{not statistically significant}. The p-value for random vs LIME, random vs attention and LIME vs attention is 0.35, 0.14 and 0.87 respectively.

This study, similar to past human-subject experiments on model simulatability, concludes that %
explanations do not \emph{definitively} help
crowdworkers to simulate text classification models.
We speculate that 
it is difficult for people
to simulate models, 
especially when they see a few fixed examples. 
A promising direction for 
future 
work could be to explore
interactive studies, 
where 
people could query the model
on inputs of their choice 
to evaluate any hypotheses they might conjecture.

\section{Limitations \& Future Directions}

There are a few important limitations of our work 
that could motivate future work in this space. %
First, our current experiments only compare explanations
that are of the same format. 
More work is required to compare 
explanations of different formats,
e.g., comparing natural language explanations 
to the top-$k\%$ highlighted tokens,
or even comparing two methods
to produce natural language explanations.
To make such comparisons, one would 
have to ensure that different explanations 
(potentially with different formats) 
communicate comparable bits of information, 
and subsequently develop learning strategies 
to train student models.

Second, validating the results 
of any automated evaluation 
with human judgement of explanation quality 
remains \emph{inherently} difficult. 
When people evaluate input attributions
(or any form of explanations) qualitatively, 
they can determine whether the attributions 
match their intuition
about what portions of the input 
should be important to solve the task 
(i.e., plausibility of explanations),
but it is not easy to evaluate if the highlighted portions
are responsible for the model's prediction.
Going forward, we think that more granular notions of simulatability, 
coupled with counterfactual access to models 
(where people can query the model) 
might help people better assess the role of explanations. 

Third, while we observe
moderate to high agreement 
amongst attribution rankings across different student architectures and learning schemes, it 
is conceivable 
that different 
explanations are favored 
based on the 
choice of student model.
This is a natural drawback of using 
a learning model for evaluation 
as the measurement could be sensitive to 
its design.
Therefore, we recommend users to average 
simulation results over a 
diverse set of student architectures, 
training examples,
and learning strategies; 
and wherever possible, 
validate explanation quality with its intended users.

Lastly, an interesting future direction
is to train explanation modules 
to generate explanations 
that optimize our metric, 
i.e., learning to produce explanations 
based on the feedback from the students.
To start with, an explanation generation module 
could be a simple transformation 
over the attention heads of the teacher model 
(as attention-based explanations 
are effective explanations as per our framework).
Learning explanations can be modeled 
as a meta-learning problem, 
where the meta-objective 
is the few-shot test performance of the student 
trained with intermediate explanations, 
and this performance could serve as a signal 
to update the explanation generation module 
using implicit gradients as in~\cite{rajeswaran2019meta}.

\section{Related Work}
\label{sec:related_work}

Several papers have suggested \emph{simulatability} 
as an approach to measure interpretability~\cite{lipton2018mythos, doshi2017towards}.
In a survey on interpretability, ~\citet{doshi2017towards}
propose the task of forward simulation: 
given an input and an explanation,
people must predict what a model would 
output for that instance. 
\citet{chandrasekaran2018explanations} 
conduct human-studies
to evaluate if explanations 
from Visual Question Answering (VQA) models 
help users predict the output.
Recently, \citet{hase-bansal-2020-evaluating}
perform a similar human-study 
across text and tabular classification tasks. 
Due to the nature of these two studies, 
the observed differences with and without explanation, 
and amongst different explanation types,
were not significant.  
Conducting large-scale human studies poses several challenges, 
including the considerable financial expense
and the logistical challenge of recruiting and retaining participants 
for unusually long tasks~\cite{chandrasekaran2018explanations}.
By automating \emph{students} in our framework, 
we mitigate such challenges, 
and observe
quantitative differences among methods in our comparisons.

Closest in spirit to our work, 
\citet{treviso2020towards} propose 
a new framework to assess explanatory power 
as the communication success rate
between an explainer and a layperson 
(which can be people or machines). 
However, as a part of their communication,
they pass on explanations during test time, 
which could easily leak the label, 
and the models trained to play this communication game 
can learn trivial protocols 
(e.g.,~explainer generating a period for positive examples 
and a comma for negative examples). 
This is probably only partially addressed 
by enforcing constraints on the explainer and the explainee. 
Our setup does not face this issue
as explanations are not available at test time.

To counter the effects of leakage due to explanations,
\citet{hase2020leakage} present a Leakage-Adjusted Simulatability (LAS) metric. 
Their metric quantifies the difference in performance 
of the simulation models (analogous to our student models) 
with and without explanations \emph{at test time}. 
To adjust for this leakage, 
they average their simulation results 
across two different sets of examples, 
ones that leak the label, and others that do not.
Leakage is modeled as a binary variable,
which is estimated by whether a discriminator 
can predict the answer using the explanation alone.
It is unclear how the average of simulation results 
solves the problem,
especially when trivial explanations leak the label.

\paragraph{Interpretability Benchmarks} 
\citet{deyoung-etal-2020-eraser} introduce ERASER benchmark 
to assess how well the rationales provided 
by models align with human rationales, 
and also how faithful these rationales are to model predictions. 
To measure faithfulness, they propose two metrics: 
comprehensiveness and sufficiency. 
They compute sufficiency by calculating 
the model performance using only the rationales, 
and comprehensiveness by measuring 
the performance without the rationales. 
This approach violates 
the i.i.d~assumption, %
as the training and evaluation data 
do not come from the same distribution. 
It is possible that 
the differences in model performance 
are due to distribution shift rather than
the features that were removed. 
This concern is also highlighted by~\citet{hooker2019benchmark}, 
who instead evaluate interpretability methods via 
their RemOve And Retrain (ROAR) benchmark.
Since the ROAR approach uses 
explanations at test time, it could be gamed:
depending upon the prediction, 
an adversarial teacher could use 
a different pre-specified ordering 
of important pixels as an explanation.
Lastly, \citet{poerner-etal-2018-evaluating} 
present a hybrid document classification task,
where the sentences are sampled 
from different documents with different class labels.
The evaluation metric validates if the important tokens 
(as per a given interpretation technique)
point to the tokens from the ``right'' document, 
i.e., one with the same label as the predicted class. 
This protocol too relies on model output 
for out-of-distribution samples (i.e., hybrid documents), 
and is very task specific.

\section{Conclusion}
\label{sec:conclusion}

We have formalized the value of explanations
as their utility
in a student-teacher framework,
measured by 
how much they improve the student's ability to simulate the teacher. 
In our setup, explanations 
are provided by the teacher as additional 
side information during training, 
but are not available at test time, thus preventing 
``leakage'' between explanations and output labels. 
Our proposed evaluation 
confirms the value 
of human-provided explanations, and correlates with  a (simplistic) notion of explanation quality.  
Additionally, we conduct extensive experiments 
that measure the value of numerous 
previously-proposed schemes for producing explanations.
Our experiments result in clear quantitative differences between different explanation methods,
which are consistent, to a moderate to high degree,
across different choices 
Among explanation methods,  we find attention to be the most effective. 
For student models, we find that both multitask and attention-regularized student learners are effective, but attention-based learners are more effective, especially in low-resource settings.

\section*{Acknowledgements}
We are grateful to Jasmijn Bastings, Katja Filippova, Matthew Lamm, Mansi Gupta, Patrick Verga, Slav Petrov and Paridhi Asija for insightful discussions that shaped this work. We also thank the TACL reviewers and action editor for providing high-quality feedback that improved our work considerably. Lastly, we acknowledge Chris Alberti for sharing explanations from the SpanBERT model.

\bibliography{emnlp2020}

\begin{thebibliography}{33}
\expandafter\ifx\csname natexlab\endcsname\relax\def\natexlab#1{#1}\fi

\bibitem[{Bahdanau et~al.(2015)Bahdanau, Cho, and Bengio}]{bahdanau2014neural}
Dzmitry Bahdanau, Kyunghyun Cho, and Yoshua Bengio. 2015.
\newblock \href {http://arxiv.org/abs/1409.0473} {Neural machine translation by
  jointly learning to align and translate}.
\newblock \emph{3rd International Conference on Learning Representations,
  {ICLR}}.

\bibitem[{Bao et~al.(2018)Bao, Chang, Yu, and
  Barzilay}]{bao-etal-2018-deriving}
Yujia Bao, Shiyu Chang, Mo~Yu, and Regina Barzilay. 2018.
\newblock \href {https://aclanthology.org/D18-1216} {Deriving machine attention
  from human rationales}.
\newblock In \emph{Proceedings of the 2018 Conference on Empirical Methods in
  Natural Language Processing}, pages 1903--1913.

\bibitem[{Chandrasekaran et~al.(2018)Chandrasekaran, Prabhu, Yadav,
  Chattopadhyay, and Parikh}]{chandrasekaran2018explanations}
Arjun Chandrasekaran, Viraj Prabhu, Deshraj Yadav, Prithvijit Chattopadhyay,
  and Devi Parikh. 2018.
\newblock \href {https://aclanthology.org/D18-1128} {Do explanations make {VQA}
  models more predictable to a human?}
\newblock In \emph{Proceedings of the 2018 Conference on Empirical Methods in
  Natural Language Processing}, pages 1036--1042.

\bibitem[{Chang et~al.(2009)Chang, Gerrish, Wang, Boyd-graber, and
  Blei}]{chang2009leaves}
Jonathan Chang, Sean Gerrish, Chong Wang, Jordan Boyd-graber, and David Blei.
  2009.
\newblock \href
  {https://proceedings.neurips.cc/paper/2009/file/f92586a25bb3145facd64ab20fd554ff-Paper.pdf}
  {Reading tea leaves: How humans interpret topic models}.
\newblock In \emph{Advances in Neural Information Processing Systems},
  volume~22.

\bibitem[{Chang et~al.(2020)Chang, Zhang, Yu, and
  Jaakkola}]{chang2020invariant}
Shiyu Chang, Yang Zhang, Mo~Yu, and Tommi~S. Jaakkola. 2020.
\newblock \href {http://proceedings.mlr.press/v119/chang20c.html} {Invariant
  rationalization}.
\newblock In \emph{Proceedings of the 37th International Conference on Machine
  Learning, {ICML} 2020, 13-18 July 2020, Virtual Event}, volume 119 of
  \emph{Proceedings of Machine Learning Research}, pages 1448--1458.

\bibitem[{Devlin et~al.(2019)Devlin, Chang, Lee, and
  Toutanova}]{devlin-etal-2019-bert}
Jacob Devlin, Ming-Wei Chang, Kenton Lee, and Kristina Toutanova. 2019.
\newblock \href {https://www.aclweb.org/anthology/N19-1423} {{BERT}:
  Pre-training of deep bidirectional transformers for language understanding}.
\newblock In \emph{Proceedings of the 2019 Conference of the North {A}merican
  Chapter of the Association for Computational Linguistics: Human Language
  Technologies, Volume 1 (Long and Short Papers)}, pages 4171--4186.

\bibitem[{DeYoung et~al.(2020)DeYoung, Jain, Rajani, Lehman, Xiong, Socher, and
  Wallace}]{deyoung-etal-2020-eraser}
Jay DeYoung, Sarthak Jain, Nazneen~Fatema Rajani, Eric Lehman, Caiming Xiong,
  Richard Socher, and Byron~C. Wallace. 2020.
\newblock \href {https://www.aclweb.org/anthology/2020.acl-main.408} {{ERASER}:
  {A} benchmark to evaluate rationalized {NLP} models}.
\newblock In \emph{Proceedings of the 58th Annual Meeting of the Association
  for Computational Linguistics}, pages 4443--4458.

\bibitem[{Dhamdhere et~al.(2019)Dhamdhere, Sundararajan, and
  Yan}]{dhamdhere2018important}
Kedar Dhamdhere, Mukund Sundararajan, and Qiqi Yan. 2019.
\newblock \href {https://openreview.net/forum?id=SylKoo0cKm} {How important is
  a neuron}.
\newblock In \emph{7th International Conference on Learning Representations,
  {ICLR}}.

\bibitem[{Doshi-Velez and Kim(2017)}]{doshi2017towards}
Finale Doshi-Velez and Been Kim. 2017.
\newblock \href {https://arxiv.org/abs/1702.08608} {Towards a rigorous science
  of interpretable machine learning}.
\newblock \emph{arXiv preprint arXiv:1702.08608}.

\bibitem[{Hase and Bansal(2020)}]{hase-bansal-2020-evaluating}
Peter Hase and Mohit Bansal. 2020.
\newblock \href {https://www.aclweb.org/anthology/2020.acl-main.491}
  {Evaluating explainable {AI}: Which algorithmic explanations help users
  predict model behavior?}
\newblock In \emph{Proceedings of the 58th Annual Meeting of the Association
  for Computational Linguistics}, pages 5540--5552.

\bibitem[{Hase et~al.(2020)Hase, Zhang, Xie, and Bansal}]{hase2020leakage}
Peter Hase, Shiyue Zhang, Harry Xie, and Mohit Bansal. 2020.
\newblock \href {https://aclanthology.org/2020.findings-emnlp.390}
  {Leakage-adjusted simulatability: Can models generate non-trivial
  explanations of their behavior in natural language?}
\newblock In \emph{Findings of the Association for Computational Linguistics:
  EMNLP 2020}, pages 4351--4367.

\bibitem[{Hooker et~al.(2019)Hooker, Erhan, Kindermans, and
  Kim}]{hooker2019benchmark}
Sara Hooker, Dumitru Erhan, Pieter-Jan Kindermans, and Been Kim. 2019.
\newblock \href
  {https://proceedings.neurips.cc/paper/2019/hash/fe4b8556000d0f0cae99daa5c5c5a410-Abstract.html}
  {A benchmark for interpretability methods in deep neural networks}.
\newblock In \emph{Advances in Neural Information Processing Systems}, pages
  9737--9748.

\bibitem[{Jacovi and Goldberg(2021)}]{jacovi2020aligning}
Alon Jacovi and Yoav Goldberg. 2021.
\newblock \href
  {https://direct.mit.edu/tacl/article-pdf/doi/10.1162/tacl\_a\_00367/1923972/tacl\_a\_00367.pdf}
  {{Aligning Faithful Interpretations with their Social Attribution}}.
\newblock \emph{Transactions of the Association for Computational Linguistics},
  9:294--310.

\bibitem[{Joshi et~al.(2020)Joshi, Chen, Liu, Weld, Zettlemoyer, and
  Levy}]{joshi2020spanbert}
Mandar Joshi, Danqi Chen, Yinhan Liu, Daniel~S Weld, Luke Zettlemoyer, and Omer
  Levy. 2020.
\newblock \href {https://transacl.org/ojs/index.php/tacl/article/view/1853}
  {Span{BERT}: Improving pre-training by representing and predicting spans}.
\newblock \emph{Transactions of the Association for Computational Linguistics},
  8:64--77.

\bibitem[{Kwiatkowski et~al.(2019)Kwiatkowski, Palomaki, Redfield, Collins,
  Parikh, Alberti, Epstein, Polosukhin, Devlin, Lee
  et~al.}]{kwiatkowski2019natural}
Tom Kwiatkowski, Jennimaria Palomaki, Olivia Redfield, Michael Collins, Ankur
  Parikh, Chris Alberti, Danielle Epstein, Illia Polosukhin, Jacob Devlin,
  Kenton Lee, et~al. 2019.
\newblock \href {https://transacl.org/ojs/index.php/tacl/article/view/1455}
  {Natural questions: a benchmark for question answering research}.
\newblock \emph{Transactions of the Association for Computational Linguistics},
  7:453--466.

\bibitem[{Lafferty et~al.(2001)Lafferty, McCallum, and
  Pereira}]{lafferty2001conditional}
John Lafferty, Andrew McCallum, and Fernando~CN Pereira. 2001.
\newblock \href {http://www.aladdin.cs.cmu.edu/papers/pdfs/y2001/crf.pdf}
  {Conditional random fields: Probabilistic models for segmenting and labeling
  sequence data}.
\newblock \emph{18th International Conference on Machine Learning 2001 (ICML
  2001)}, pages 282--289.

\bibitem[{Lamm et~al.(2021)Lamm, Palomaki, Alberti, Andor, Choi, Soares, and
  Collins}]{lamm2020qed}
Matthew Lamm, Jennimaria Palomaki, Chris Alberti, Daniel Andor, Eunsol Choi,
  Livio~Baldini Soares, and Michael Collins. 2021.
\newblock \href
  {https://direct.mit.edu/tacl/article-pdf/doi/10.1162/tacl\_a\_00398/1955181/tacl\_a\_00398.pdf}
  {{QED: A Framework and Dataset for Explanations in Question Answering}}.
\newblock \emph{Transactions of the Association for Computational Linguistics},
  9:790--806.

\bibitem[{Lipton(2016)}]{lipton2018mythos}
Zachary~C Lipton. 2016.
\newblock \href {http://doi.acm.org/10.1145/3236386.3241340} {The mythos of
  model interpretability}.
\newblock \emph{{ACM} Queue}, 16(3):31--57.

\bibitem[{Maas et~al.(2011)Maas, Daly, Pham, Huang, Ng, and
  Potts}]{maas-EtAl:2011:ACL-HLT2011}
Andrew~L. Maas, Raymond~E. Daly, Peter~T. Pham, Dan Huang, Andrew~Y. Ng, and
  Christopher Potts. 2011.
\newblock \href {http://www.aclweb.org/anthology/P11-1015} {Learning word
  vectors for sentiment analysis}.
\newblock In \emph{Proceedings of the 49th Annual Meeting of the Association
  for Computational Linguistics: Human Language Technologies}, pages 142--150.

\bibitem[{Mercier and Sperber(2017)}]{mercier2017enigma}
Hugo Mercier and Dan Sperber. 2017.
\newblock \href {https://www.hup.harvard.edu/catalog.php?isbn=9780674237827}
  {\emph{The enigma of reason}}.
\newblock Harvard University Press.

\bibitem[{Poerner et~al.(2018)Poerner, Sch{\"u}tze, and
  Roth}]{poerner-etal-2018-evaluating}
Nina Poerner, Hinrich Sch{\"u}tze, and Benjamin Roth. 2018.
\newblock \href {https://www.aclweb.org/anthology/P18-1032} {Evaluating neural
  network explanation methods using hybrid documents and morphosyntactic
  agreement}.
\newblock In \emph{Proceedings of the 56th Annual Meeting of the Association
  for Computational Linguistics (Volume 1: Long Papers)}, pages 340--350.

\bibitem[{Pruthi et~al.(2020)Pruthi, Dhingra, Neubig, and
  Lipton}]{pruthi2020weakly}
Danish Pruthi, Bhuwan Dhingra, Graham Neubig, and Zachary~C Lipton. 2020.
\newblock \href {https://aclanthology.org/2020.findings-emnlp.353} {Weakly-and
  semi-supervised evidence extraction}.
\newblock \emph{Findings of the Association for Computational Linguistics:
  {EMNLP} 2020}, pages 3965--3970.

\bibitem[{Rajeswaran et~al.(2019)Rajeswaran, Finn, Kakade, and
  Levine}]{rajeswaran2019meta}
Aravind Rajeswaran, Chelsea Finn, Sham~M. Kakade, and Sergey Levine. 2019.
\newblock \href
  {https://proceedings.neurips.cc/paper/2019/hash/072b030ba126b2f4b2374f342be9ed44-Abstract.html}
  {Meta-learning with implicit gradients}.
\newblock In \emph{Advances in Neural Information Processing Systems 32: Annual
  Conference on Neural Information Processing Systems}, pages 113--124.

\bibitem[{Ribeiro et~al.(2016)Ribeiro, Singh, and Guestrin}]{ribeiro2016should}
Marco~Tulio Ribeiro, Sameer Singh, and Carlos Guestrin. 2016.
\newblock \href {https://doi.org/10.1145/2939672.2939778} {"why should {I}
  trust you?": Explaining the predictions of any classifier}.
\newblock In \emph{Proceedings of the 22nd ACM SIGKDD international conference
  on knowledge discovery and data mining}, pages 1135--1144.

\bibitem[{Shrikumar et~al.(2017)Shrikumar, Greenside, and
  Kundaje}]{shrikumar2017learning}
Avanti Shrikumar, Peyton Greenside, and Anshul Kundaje. 2017.
\newblock \href {http://proceedings.mlr.press/v70/shrikumar17a.html} {Learning
  important features through propagating activation differences}.
\newblock In \emph{Proceedings of the 34th International Conference on Machine
  Learning, {ICML} 2017, Sydney, NSW, Australia, 6-11 August 2017}, volume~70
  of \emph{Proceedings of Machine Learning Research}, pages 3145--3153.

\bibitem[{Simonyan et~al.(2014)Simonyan, Vedaldi, and
  Zisserman}]{simonyan2013deep}
Karen Simonyan, Andrea Vedaldi, and Andrew Zisserman. 2014.
\newblock \href {http://arxiv.org/abs/1312.6034} {Deep inside convolutional
  networks: Visualising image classification models and saliency maps}.
\newblock In \emph{2nd International Conference on Learning Representations,
  {ICLR}}.

\bibitem[{Sundararajan et~al.(2017)Sundararajan, Taly, and
  Yan}]{sundararajan2017axiomatic}
Mukund Sundararajan, Ankur Taly, and Qiqi Yan. 2017.
\newblock \href {http://proceedings.mlr.press/v70/sundararajan17a.html}
  {Axiomatic attribution for deep networks}.
\newblock In \emph{Proceedings of the 34th International Conference on Machine
  Learning, {ICML}}, volume~70 of \emph{Proceedings of Machine Learning
  Research}, pages 3319--3328.

\bibitem[{Treviso and Martins(2020)}]{treviso2020towards}
Marcos~V. Treviso and Andr{\'{e}} F.~T. Martins. 2020.
\newblock \href {https://doi.org/10.18653/v1/2020.blackboxnlp-1.10} {The
  explanation game: Towards prediction explainability through sparse
  communication}.
\newblock In \emph{Proceedings of the Third BlackboxNLP Workshop on Analyzing
  and Interpreting Neural Networks for NLP, BlackboxNLP@EMNLP 2020, Online,
  November 2020}, pages 107--118.

\bibitem[{Vaswani et~al.(2017)Vaswani, Shazeer, Parmar, Uszkoreit, Jones,
  Gomez, Kaiser, and Polosukhin}]{vaswani2017attention}
Ashish Vaswani, Noam Shazeer, Niki Parmar, Jakob Uszkoreit, Llion Jones,
  Aidan~N Gomez, {\L}ukasz Kaiser, and Illia Polosukhin. 2017.
\newblock \href
  {https://proceedings.neurips.cc/paper/2017/hash/3f5ee243547dee91fbd053c1c4a845aa-Abstract.html}
  {Attention is all you need}.
\newblock In \emph{Advances in neural information processing systems}, pages
  5998--6008.

\bibitem[{Yin et~al.(2021)Yin, Fernandes, Pruthi, Chaudhary, Martins, and
  Neubig}]{yin21acl}
Kayo Yin, Patrick Fernandes, Danish Pruthi, Aditi Chaudhary, André F.~T.
  Martins, and Graham Neubig. 2021.
\newblock \href {https://arxiv.org/abs/2105.06977} {Do context-aware
  translation models pay the right attention?}
\newblock In \emph{Joint Conference of the 59th Annual Meeting of the
  Association for Computational Linguistics and the 11th International Joint
  Conference on Natural Language Processing (ACL-IJCNLP)}, Virtual.

\bibitem[{Zaidan and Eisner(2008)}]{zaidan-eisner-2008-modeling}
Omar Zaidan and Jason Eisner. 2008.
\newblock \href {https://www.aclweb.org/anthology/D08-1004} {Modeling
  annotators: {A} generative approach to learning from annotator rationales}.
\newblock In \emph{Proceedings of the 2008 Conference on Empirical Methods in
  Natural Language Processing}, pages 31--40.

\bibitem[{Zaidan et~al.(2007)Zaidan, Eisner, and Piatko}]{zaidan2007using}
Omar Zaidan, Jason Eisner, and Christine Piatko. 2007.
\newblock \href {https://aclanthology.org/N07-1033/} {Using “annotator
  rationales” to improve machine learning for text categorization}.
\newblock In \emph{Human language technologies 2007: The conference of the
  North American chapter of the association for computational linguistics;
  proceedings of the main conference}, pages 260--267.

\bibitem[{Zhong et~al.(2019)Zhong, Shao, and McKeown}]{zhong2019fine}
Ruiqi Zhong, Steven Shao, and Kathleen McKeown. 2019.
\newblock \href {http://arxiv.org/abs/1908.06870} {Fine-grained sentiment
  analysis with faithful attention}.
\newblock \emph{arXiv preprint arXiv:1908.06870}.

\end{thebibliography}
\clearpage

\begin{table*}
\centering
\footnotesize
\begin{tabular}{
>{\columncolor[HTML]{FFFFFF}}c 
>{\columncolor[HTML]{EFEFEF}}c ccccccccc}
\toprule
\multicolumn{1}{l}{\cellcolor[HTML]{FFFFFF}} &
  \multicolumn{1}{l}{\cellcolor[HTML]{FFFFFF}} &
  \multicolumn{2}{c}{\cellcolor[HTML]{FFFFFF}BERT Large} &
  \multicolumn{2}{c}{\cellcolor[HTML]{FFFFFF}BERT Base} &
  \multicolumn{2}{c}{\cellcolor[HTML]{FFFFFF}BiLSTM Attn.} &
  \cellcolor[HTML]{FFFFFF}BiLSTM &
  \multicolumn{2}{c}{\cellcolor[HTML]{FFFFFF}ERASER} \\
\multicolumn{1}{l}{\cellcolor[HTML]{FFFFFF}} &
  \multicolumn{1}{l}{\cellcolor[HTML]{EFEFEF}\textbf{}} & 
  \cellcolor[HTML]{EFEFEF}MTL &
  \cellcolor[HTML]{EFEFEF}AR &
  \cellcolor[HTML]{EFEFEF}MTL &
  \cellcolor[HTML]{EFEFEF}AR &
  \cellcolor[HTML]{EFEFEF}MTL &
  \cellcolor[HTML]{EFEFEF}AR &
  \cellcolor[HTML]{EFEFEF}MTL &
  \cellcolor[HTML]{EFEFEF}Suffic. &
  \cellcolor[HTML]{EFEFEF}Compr. \\
\cellcolor[HTML]{FFFFFF} &
  MTL &
  \cellcolor[HTML]{BCD4E6}1.00 &
  \cellcolor[HTML]{D1E2EE}0.69 &
  \cellcolor[HTML]{DFEBF3}0.49 &
  \cellcolor[HTML]{E0EBF4}0.47 &
  \cellcolor[HTML]{DAE7F1}0.57 &
  \cellcolor[HTML]{E3EDF5}0.42 &
  \cellcolor[HTML]{EDF3F8}0.28 &
  \cellcolor[HTML]{F0F6FA}0.23 &
  \cellcolor[HTML]{E5EEF5}0.40 \\
\multirow{-2}{*}{\cellcolor[HTML]{FFFFFF}\begin{tabular}[c]{@{}c@{}}BERT \\ Large\end{tabular}} &
  AR &
  \cellcolor[HTML]{D1E2EE}0.69 &
  \cellcolor[HTML]{BCD4E6}1.00 &
  \cellcolor[HTML]{E1ECF4}0.45 &
  \cellcolor[HTML]{D1E2EE}0.69 &
  \cellcolor[HTML]{CBDEEC}0.78 &
  \cellcolor[HTML]{D4E4EF}0.64 &
  \cellcolor[HTML]{EAF2F7}0.32 &
  \cellcolor[HTML]{FEFCFA}-0.06 &
  \cellcolor[HTML]{FFFFFF}0.00 \\
  \cline{0-1}
\cellcolor[HTML]{FFFFFF} &
  MTL &
  \cellcolor[HTML]{DFEBF3}0.49 &
  \cellcolor[HTML]{E1ECF4}0.45 &
  \cellcolor[HTML]{BCD4E6}1.00 &
  \cellcolor[HTML]{E8F0F7}0.35 &
  \cellcolor[HTML]{DDE9F3}0.52 &
  \cellcolor[HTML]{E3EDF5}0.42 &
  \cellcolor[HTML]{E3EDF5}0.42 &
  \cellcolor[HTML]{E2EDF5}0.44 &
  \cellcolor[HTML]{E2EDF5}0.44 \\
\multirow{-2}{*}{\cellcolor[HTML]{FFFFFF}\begin{tabular}[c]{@{}c@{}}BERT \\ Base\end{tabular}} &
  AR &
  \cellcolor[HTML]{E0EBF4}0.47 &
  \cellcolor[HTML]{D1E2EE}0.69 &
  \cellcolor[HTML]{E8F0F7}0.35 &
  \cellcolor[HTML]{BCD4E6}1.00 &
  \cellcolor[HTML]{DDE9F3}0.52 &
  \cellcolor[HTML]{D6E5F0}0.61 &
  \cellcolor[HTML]{ECF3F8}0.30 &
  \cellcolor[HTML]{EFF5F9}0.25 &
  \cellcolor[HTML]{DCE8F2}0.54 \\
  \cline{0-1}
\cellcolor[HTML]{FFFFFF} &
  MTL &
  \cellcolor[HTML]{DAE7F1}0.57 &
  \cellcolor[HTML]{CBDEEC}0.78 &
  \cellcolor[HTML]{DDE9F3}0.52 &
  \cellcolor[HTML]{DDE9F3}0.52 &
  \cellcolor[HTML]{BCD4E6}1.00 &
  \cellcolor[HTML]{C4D9E9}0.89 &
  \cellcolor[HTML]{D9E7F1}0.57 &
  \cellcolor[HTML]{E7F0F6}0.37 &
  \cellcolor[HTML]{D4E4EF}0.65 \\
\multirow{-2}{*}{\cellcolor[HTML]{FFFFFF}\begin{tabular}[c]{@{}c@{}}BiLSTM \\ Attn.\end{tabular}} &
  AR &
  \cellcolor[HTML]{E3EDF5}0.42 &
  \cellcolor[HTML]{D4E4EF}0.64 &
  \cellcolor[HTML]{E3EDF5}0.42 &
  \cellcolor[HTML]{D6E5F0}0.61 &
  \cellcolor[HTML]{C4D9E9}0.89 &
  \cellcolor[HTML]{BCD4E6}1.00 &
  \cellcolor[HTML]{D6E5F0}0.61 &
  \cellcolor[HTML]{E3EDF5}0.42 &
  \cellcolor[HTML]{D8E6F1}0.59 \\
  \cline{0-1}
BiLSTM &
  MTL &
  \cellcolor[HTML]{EDF3F8}0.28 &
  \cellcolor[HTML]{EAF2F7}0.32 &
  \cellcolor[HTML]{E3EDF5}0.42 &
  \cellcolor[HTML]{ECF3F8}0.30 &
  \cellcolor[HTML]{D9E7F1}0.57 &
  \cellcolor[HTML]{D6E5F0}0.61 &
  \cellcolor[HTML]{BCD4E6}1.00 &
  \cellcolor[HTML]{CDDFEC}0.76 &
  \cellcolor[HTML]{DFEBF4}0.48 \\
  \cline{0-1}
\cellcolor[HTML]{FFFFFF} &
  Suffic. &
  \cellcolor[HTML]{F0F6FA}0.23 &
  \cellcolor[HTML]{FEFCFA}-0.06 &
  \cellcolor[HTML]{E2EDF5}0.44 &
  \cellcolor[HTML]{EFF5F9}0.25 &
  \cellcolor[HTML]{E7F0F6}0.37 &
  \cellcolor[HTML]{E3EDF5}0.42 &
  \cellcolor[HTML]{CDDFEC}0.76 &
  \cellcolor[HTML]{BCD4E6}1.00 &
  \cellcolor[HTML]{D7E5F0}0.61 \\
\multirow{-2}{*}{\cellcolor[HTML]{FFFFFF}ERASER} &
  Compr. &
  \cellcolor[HTML]{E5EEF5}0.40 &
  \cellcolor[HTML]{FFFFFF}0.00 &
  \cellcolor[HTML]{E2EDF5}0.44 &
  \cellcolor[HTML]{DCE8F2}0.54 &
  \cellcolor[HTML]{D4E4EF}0.65 &
  \cellcolor[HTML]{D8E6F1}0.59 &
  \cellcolor[HTML]{DFEBF4}0.48 &
  \cellcolor[HTML]{D7E5F0}0.61 &
  \cellcolor[HTML]{BCD4E6}1.00 \\ \bottomrule
\end{tabular}
\caption{The Kendall rank correlation coefficient, $\tau$, comparing rankings obtained through different settings of our metric. We also compute correlations with the sufficiency and comprehensiveness metrics from the ERASER benchmark~\cite{deyoung-etal-2020-eraser}.  MTL and AR denote Multitask Learning and Attention Regularization. Values can range from \hln{-1.0} (perfect disagreement) to \hlp{1.0} (perfect agreement). Across different students and different learning strategies, the rankings obtained are highly correlated. }
\label{tbl:kendall_tau_coefficient}
\end{table*}

\begin{table*}[t]
    \centering
    \begin{tabular}{lrrrrlrrrr}
    \toprule
                         & \multicolumn{4}{c}{Attention Regularization} &  & \multicolumn{4}{c}{Multitask Learning} \\
                            \cline{2-5} \cline {7-10}
    Value of $k$             & 5\%     & 10\%    & 20\%   & 40\%   &  & 5\%           & 10\% & 20\% & 40\%  \\ \midrule
    LIME       & 93.0    & 92.6   & 92.5   & 92.0   &  & 92.8 & 92.6 & 92.5 & 91.8 \\
    Gradient Norm   & 92.8 & 92.4 & 90.6 & 90.6 & & 93.1 & 93.1 & 92.9 & 93.0 \\ 
    Gradient $\times$ Input       &92.5	 & 92.2 &	92.6 & 92.8   &  & 92.4 & 92.7 &		92.5 & 91.3 \\ 
    Layer Conductance       & 93.6 & 93.5 & 93.4 &  92.9 &  & 92.2	& 92.9 &	92.5	& 92.3 \\
    Integrated Gradients       & 94.1&	93.6&	93.6&	93.1  &  & 93.4	& 93.3 &	93.1	& 92.1 \\
    Attention & \textbf{94.7} &  \textbf{95.2} & \textbf{95.3} & \textbf{94.6} &  & \textbf{94.0} & \textbf{94.4} & \textbf{94.7}  & \textbf{94.9} \\
    \bottomrule
    \end{tabular}
    \caption{Simulation accuracy of a BERT-base student model, examining the effect of $k$ in selecting top-$k$\% explanatory tokens.  Student model without explanations obtains a simulation accuracy of $92.6$.}
    \label{tbl:results_movie_reviews_varying_k}
\end{table*}

\section*{Supplementary Material}
\section{Explanation Types}
\label{sec:appendix_exp}

We examine the following attribution methods:
\paragraph{LIME} Locally Interpretable Model-agnostic Explanations~\cite{ribeiro2016should}, or LIME, 
are explanations produced by a linear interpretable 
model that is trained 
to approximate 
the original black box 
model in the local neighborhood of the input example.
For a given example, several samples are constructed by 
perturbing the input string, and these samples 
are used to train the
linear model. 
We draw twice as many samples 
as the number of tokens in the example,
and select the top words that explain the predicted class. 
We set the number of features for the linear classifier to be $2k$, where $k$ is the number of tokens to be selected. 

\paragraph{Gradient-based Saliency Methods} Several papers,
both in NLP and computer vision,
use gradients of the log-likelihood of the predicted label
to 
understand the effect of infinitesimally small perturbations
in the input. 
While no 
perturbation of an input string is 
infinitesimally small, nonetheless, researchers have continued 
to use this metric. It is most commonly used in two forms:
grad norm, i.e., the $\ell_2$ norm of the gradient w.r.t. the token representation,
and grad $\times$ input (also called grad dot), i.e., the dot product of the gradient w.r.t the token representation and the token representation. 

\paragraph{Integrated Gradients} Gradients capture 
only the effect of perturbations 
in an infinitesimally small neighborhood, integrated gradients~\cite{sundararajan2017axiomatic},
instead compute and integrate 
gradients 
along the line
joining a starting reference point and the given input example. 
For each example, we integrate the gradients over $50$ points on the line. 

\paragraph{Layer Conductance} \citet{dhamdhere2018important} introduce and extend the notion of \textit{conductance} to compute neuron-level importance scores. We apply layer conductance on the first encoder layer of our teacher model and aggregate the scores to define the attributions over the input tokens. 

\paragraph{DeepLIFT}
DeepLIFT uses a reference 
for the model input and target, and measures the contribution of each input feature in the pair-wise difference from this reference~\cite{shrikumar2017learning}. It addresses the limitations of gradient-based attribution methods, for regimes with zero and discontinuous gradients. It backpropagates the contributions of neurons using multipliers as in partial derivatives. We use a reference input (embeddings) of all zeros for our experiments.

\paragraph{Attention-based Explanations}

Attention mechanisms were originally introduced by~\citet{bahdanau2014neural} 
to align source and target tokens in neural machine translation.
Because  attention mechanisms allocate weight among the encoded tokens,
these coefficients are sometimes thought of intuitively
as indicating which tokens the model \emph{focuses on} 
when making a prediction.

\clearpage
\appendix

\bibliographystyle{acl_natbib}

\end{document}